\documentclass[a4paper,11pt]{report}
\usepackage{algorithm2e}
\usepackage{amsmath}
\usepackage{amssymb}
\usepackage{amsthm}
\usepackage{array}
\usepackage[english]{babel}
\usepackage{bbm}
\usepackage{bm}
\usepackage{fancyhdr}
\usepackage{float}
\usepackage{geometry}
\usepackage{graphicx}
\usepackage{hyperref}
\usepackage[utf8]{inputenc}
\usepackage{listings}
\usepackage[authoryear,round]{natbib}
\usepackage[intoc]{nomencl}
\usepackage[nottoc]{tocbibind}
\usepackage{xcolor}

\newcounter{lemma}
\newcounter{theorem}
\numberwithin{theorem}{chapter}
\numberwithin{lemma}{chapter}
\newtheorem{lemma_technical}[lemma]{Lemma}
\newtheorem{theorem_exp3_unifkminus1_util}[theorem]{Theorem}
\newtheorem{theorem_exp3_unifkminus1_nonutil}[theorem]{Theorem}

\makeatletter
\newcommand{\algotitle}[2]{%
  \stepcounter{algocf}%
  \hypertarget{algocf.title.\theHalgocf}{}%
  \NR@gettitle{#1}%
  \label{#2}%
  \addtocounter{algocf}{-1}%
}
\makeatother

\makenomenclature

\newcommand{\lta}{\widetilde{\ell}_t^a}
\newcommand{\lti}{\widetilde{\ell}_t^i}
\newcommand{\histfull}[1]{(A_1, B_1, Y_1), \hdots, (A_{#1}, B_{#1}, Y_{#1})}
\newcommand{\hist}[1]{\mathcal H_{#1}}

\title{Duelling Bandits with Weak Regret in Adversarial Environments}
\date{August 6, 2018}

\begin{document}
\makeatletter
\begin{titlepage}
  \centering
  {\scshape \LARGE University of Copenhagen \par}
  \vspace{1cm}
  {\scshape \LARGE Master's Thesis \par}
  \vspace{1.5cm}
  \rule[6mm]{\textwidth}{1.5pt}
  {\huge\bfseries \@title}
  \rule[-2mm]{\textwidth}{1.5pt}
  \setlength{\tabcolsep}{1.5cm}
  \Large
  \vskip 1cm
  \begin{tabular}{lr}
    \textit{Author:} & \textit{Supervisor:} \\
    Lennard {\scshape Hilgendorf} & Yevgeny {\scshape Seldin} \\
    fwc817@alumni.ku.dk & seldin@di.ku.dk
  \end{tabular}
  \vskip 10cm
  \@date
\end{titlepage}
\makeatother

\begin{abstract}
  Research on the multi-armed bandit problem has studied the trade-off of
  exploration and exploitation in depth. However, there are numerous
  applications where the cardinal absolute-valued feedback model (e.g.
  ratings from one to five) is not
  suitable. This has motivated the formulation of the duelling bandits
  problem,
  where the learner picks a pair of actions and observes a noisy binary
  feedback, indicating a relative preference between the two.
  There exist a multitude of different settings and interpretations of the
  problem for two
  reasons. First, due to the absence of a total order of actions, there
  is no natural definition of the best action. Existing work either
  explicitly assumes the existence of a linear order, or uses a custom
  definition for the winner. Second, there are multiple reasonable notions
  of regret to measure the learner's performance. Most prior work has
  been focussing on the \textit{strong regret}, which averages the quality
  of the two actions picked. This work focusses on the \textit{weak regret},
  which is based on the quality of the better of the two actions selected.
  Weak regret is the more appropriate performance measure when the pair's
  inferior action has no significant detrimental effect on the pair's
  quality.

  We study the duelling bandits problem in the adversarial setting.
  We provide an algorithm which has theoretical guarantees in both the
  utility-based setting, which implies a total order, and the unrestricted
  setting. For the latter, we work with the \textit{Borda winner}, finding
  the action maximising the probability of winning against an action sampled
  uniformly at random. The thesis concludes with experimental results based
  on both real-world data and synthetic data, showing the algorithm's
  performance and limitations. 
\end{abstract}

\tableofcontents

\printnomenclature
\chapter*{Notation}
\addcontentsline{toc}{chapter}{Notation}

\begin{table}[h]
  \centering
  \begin{tabular}{l|l}
    Symbol & Definition \\ \hline
    \(\mathbb E \left[ \cdot \right]\) & Expectation of random variable \\
    \(\mathbbm 1 \left( \cdot \right)\) & Indicator function \\
    \(\Delta^K\) & \(K\)-simplex \\
    \([K]\) & Set \(\left\{ 1, \hdots, K \right\}\) \\
    \(K\) & Number of arms \\
    \(T\) & Time horizon \\
    \(\mathbf M^t\) & Outcome matrix in round \(t\) \\
    \(\mathcal M\) & Sequence of outcome matrices \\
    \(\bm \ell_t\) & Loss vector in round \(t\) \\
    \(\mathcal L \) & Sequence of loss vectors \\
    \(A_t, B_t\) & Pair of actions played in round \(t\) \\
    \(Y_t\) & Outcome observed in round \(t\) \\
    \(\psi(\cdot, \cdot)\) & Loss-combining function \\
    \(R(T)\) & Cumulative regret until round \(T\) \\
    \(\phi(\cdot, \cdot)\) & Link function \\
    \(\mathbf x\) & Reward / utility vector \\
    \(a_U^*\) & Utility-based winner \\
    \(a_B^*\) & Borda winner \\
    \(a_C^*\) & Copeland winner \\
    \(\mathbf u\) & Von-Neumann winner \\
    \(R_S(T)\) & Cumulative strong regret \\
    \(R_W(T)\) & Cumulative weak regret \\
    \(R_W^U(T)\) & Utility-based cumulative weak regret \\
    \(R_W^N(T)\) & Non-utility-based cumulative weak regret \\
    \(\eta\) & Learning rate \\
    \(\widetilde \ell^a\) & Loss estimator \\
    \(\widetilde L_t(a)\) & Cumulative loss estimator \\
    \(\hist t\) & Set \(\histfull t\) \\
    \(\bar{\bm \ell}_t\) & Utility-based loss vector in round \(t\)
  \end{tabular}
\end{table}

\chapter{Introduction}
\label{ch:introduction}

The trade-off between exploration and exploitation, which arises in various
sequential decision problems and online learning problems such as
reinforcement learning, has been studied in-depth by research on the
multi-armed bandit problem. Given a set of actions, which are also termed
arms, the environment assigns a bounded real-valued utility to each of them.
Following a sequential game protocol, the learner picks an arm and observes
its noisy real-valued feedback, which is based on the arm's utility. The
learner's goal is to maximise the cumulative reward. This theoretical
framework can be applied to various practical settings where cardinal
feedback is readily available. 

However, often other feedback models are required. In the context of online
ranker
evaluation, \cite{Radlinski:2008:CDR:1458082.1458092} examined the relation
between various absolute usage metrics and the quality of retrieval
functions. They concluded that none of the measures covered were a reliable
predictor for the retrieval quality. Instead, their results suggested that
relative feedback obtained through pairwise comparisons, such as ``option A
is preferred over option B'', can be used for consistent and more accurate
estimates. Often this form of feedback is easier to obtain, making it desirable
to have algorithms which are capable of handling this learning task.

Research addressing the \textit{duelling bandits problem} deals with
formalisation of online learning problems involving pairwise comparisons
and studies algorithms selecting a sequence of pairs of actions, assuming a
binary feedback mechanism. The term has been coined by
\cite{Yue:2009:IOI:1553374.1553527}, who were motivated by the results of
the experimental studies by \cite{Radlinski:2008:CDR:1458082.1458092}.

Unlike the classical multi-armed bandit problem, the existence of a linear
order is not guaranteed, as violations of transitivity (\(A \succ B \succ C
\succ A\)) cannot be ruled out. This makes the definition of a winner
ambiguous. The first papers either explicitly assumed a linear order, or
used a convex utility-function, which in turn induces a total order.
\cite{Yue:2009:IOI:1553374.1553527} modelled the space of actions \(
\mathcal W \) as a convex subset of the \(d\)-dimensional Euclidean space,
allowing the embedding of the  parameterisation of complex retrieval
functions.
Assuming a convex function mapping actions to real-valued utilities,
outcomes of comparisons between two actions are modelled as independent
Bernoulli random variables with bias given by an odd link
function mapping the signed difference of two utilities to the zero-one
interval. In order to measure the performance of an algorithm, they
introduced a notion of regret, which corresponds to what subsequent
literature terms as \textit{strong regret} \citep{Yue:2012:KDB:2240304.2240501}.  Given a
finite time horizon \(T\), strong regret relates to the cumulative
difference
between the average quality of the pair of actions chosen at time \(t\) and
the quality of the best action in hindsight \(a^*\), which has maximum
utility. This measure reflects the relative number of users who would have
chosen the best action over an action picked uniformly at random from the
pair of actions \citep{Yue:2012:KDB:2240304.2240501}, and is zero in this setting if and only
if both actions selected are the best action.

Instead of considering an infinite space of parameterisations, it is often
more sensible to consider a more limited set of actions. Presuming that this
set of actions is finite, i.e. \( \left\lvert \mathcal W \right\rvert = K
\), \cite{Yue:2012:KDB:2240304.2240501} examine the \textit{\(K\)-armed duelling bandits
problem}. Instead of relying on a utility function, they assumed the
existence of a total order of the set of actions. Complementing the
previously defined strong regret, they introduced the notion of \textit{weak
regret}, which reflects the relative number of users preferring the overall
best arm over the better action of the pair of actions presented. To be
zero, it suffices that the better action is identical to the best action.

In practice, the assumption of a total order is often violated, limiting the
works' applicability \citep{pmlr-v40-Dudik15}. Some works assume the
existence of a \textit{Condorcet winner}, which is an action winning against
all other actions with probability \(> \frac 1 2\).
\cite{Urvoy:2013:GEK:3042817.3042904} dropped the assumption of utilities
and linear orders altogether, giving rise to the stochastic
\textit{non-utility-based} duelling bandits problem. Recognising that this
relaxation might impede the existence of a Condorcet winner, they resort to
the corpus of social choice theory and voting theory, basing the definition
of the winner and the mechanism inducing regret on the \textit{Borda score},
which relates to the probability of a particular action winning against an
action sampled uniformly at random, as well as on the \textit{Copeland
score}, which takes into account the number of other actions an action is
preferred to. The latter has been covered in greater extent by 
\cite{Zoghi:2015:CDB:2969239.2969274} as a natural generalisation of the
Condorcet winner. A game-theoretic interpretation of the duelling bandits
problem was initially suggested by \cite{pmlr-v40-Dudik15}. In an attempt to
address certain shortcomings of the Borda winner and the Copeland winner,
they introduced the notion of the \textit{von-Neumann winner}, which is a
distribution over actions which beats every other policy with
probability \(\geq \frac 1 2\). In addition, they discussed the case where
outcomes are no longer sampled from a stationary distribution, but instead
are generated from a distribution which is selected by an adversary in an
arbitrary way on a per-round basis.

The adversarial duelling bandits problem was further studied by
\cite{Gajane:2015:REW:3045118.3045143}, who assumed a binary utility vector,
effectively splitting the set of actions into a set of ``good'' actions and
a set of ``bad'' actions, the former winning all duels against the latter
with probability one.

Finally, \cite{pmlr-v70-chen17c} revisited the notion of weak regret by
\cite{Yue:2012:KDB:2240304.2240501}, proving bounds on the expected regret constant in \(T\),
assuming a Condorcet winner.

\begin{table}[h]
  \renewcommand{\thefootnote}{\fnsymbol{footnote}}
  \renewcommand{\thempfootnote}{\fnsymbol{mpfootnote}}
  \newcommand{\strong}{\footnotemark[2]}
  \newcommand{\weak}{\footnotemark[4]}
  \newcommand{\cstrong}[1]{\textcolor{black}{#1} \strong}
  \newcommand{\cweak}[1]{\textcolor{blue}{#1} \weak}
  \newcommand{\cboth}[1]{\textcolor{blue}{#1} \strong \weak}
  \begin{minipage}{1.0\textwidth}
    \centering
    \begin{tabular}{m{3.9cm}||m{4.7cm}|m{3.5cm}}
      & Stochastic Setting & Adversarial Setting \\ \hline \hline
      Utility-based setting &
      \shortstack[l]{
        \cstrong{\cite{Yue:2009:IOI:1553374.1553527}} \\
        \cboth{\cite{Yue:2012:KDB:2240304.2240501}} \\
        \cstrong{\cite{Ailon:2014:RDB:3044805.3044988}} \\
        \cweak{\cite{DBLP:journals/corr/ChenF16}} \\
        \cboth{\cite{pmlr-v70-chen17c}}
      } & \shortstack[l]{
        \cstrong{\cite{Gajane:2015:REW:3045118.3045143}} \\
        \cweak{\textbf{This work}}
      } \\ \hline

      Condorcet winner &
      \shortstack[l]{
        \cstrong{\cite{Yue:2011:BMB:3104482.3104513}} \\
        \cstrong{\cite{Urvoy:2013:GEK:3042817.3042904}} \\
        \cstrong{\cite{Zoghi:2014:RUC:3044805.3044894}} \\
        \cstrong{\cite{pmlr-v40-Komiyama15}} \\
        \cboth{\cite{pmlr-v70-chen17c}}
      } & \\ \hline

      Borda winner &
      \shortstack[l]{
        \cstrong{\cite{Urvoy:2013:GEK:3042817.3042904}} \\
        \textcolor{red}{\cite{pmlr-v38-jamieson15}} \footnotemark[1]
      } & \shortstack[l]{
        \cweak{\textbf{This work}}
      } \\ \hline

      Copeland winner &
      \shortstack[l]{
        \cstrong{\cite{Urvoy:2013:GEK:3042817.3042904}} \\
        \cstrong{\cite{Zoghi:2015:CDB:2969239.2969274}} \\
        \cstrong{\cite{NIPS2016_6157}}
      } & \\ \hline
      
      Von-Neumann winner & 
      \shortstack[l]{
        \cstrong{\cite{pmlr-v49-balsubramani16}}
      } & \shortstack[l]{
        \cstrong{\cite{pmlr-v40-Dudik15}}
      }
    \end{tabular}
    \caption{Overview of existing work}
    \label{tab:prior_work_summary}
    \footnotetext[2]{\textcolor{black}{Studied in strong regret setting}}
    \footnotetext[4]{\textcolor{blue}{Studied in weak regret setting}}
    \footnotetext[1]{\textcolor{red}{Pure exploration, bounded number of
    rounds until termination for best arm identification}}
  \end{minipage}
\end{table}

As shown in the summary in Table~\ref{tab:prior_work_summary}, 
most prior work has been focussing on the strong regret to evaluate the
quality of pairs of actions. The suitability of the strong regret
depends on the application and
assumption made: if the presence of undesired actions has a negative impact
on the perceived quality or if the user's experience can be enhanced by
showing two good options, e.g. when considering search results
\citep{pmlr-v70-chen17c}, strong regret is a reasonable modelling
assumption. However, if the quality of pairs consisting of a good and a bad
action
is dominated by the quality of the former, it might be more appropriate to
consider the weak regret. At the same time, the requirement that outcomes
are sampled from a stationary distribution might not always be adequate in
practice. To our knowledge, algorithms for weak regret have only been
devised in the stochastic setting requiring a Condorcet winner. So far
studies of the adversarial duelling bandits problem have been limited to the
von-Neumann setting and, with with some limitations, the utility-based
setting, limiting its practical applicability.

This work provides a framework consolidating the different problem
formulations, facilitating discussion of results by
relating the utility-based setting to the non-utility-based setting.
Subsequently, we discuss the suitability of a range of algorithms assuming
different winner models and support our claims experimentally.  Finally, we
propose an algorithm covering both the utility-based setting and the
non-utility-based setting of the adversarial duelling bandits problem, the
latter using the Borda winner to define the winner and quality of individual
actions. We restrict ourselves to weak regret, proving an upper bound on the
expected regret of \( O \left( \sqrt{K T \log K} \right) \), extending the
work on adversarial duelling bandits by \cite{pmlr-v40-Dudik15} and
\cite{Gajane:2015:REW:3045118.3045143}, covering a different winner setting
and regret measure. Our contribution advances the applicability of duelling
bandits algorithms by providing a robust algorithm with explicit regret
guarantees in adversarial settings.

The remainder of this thesis is structured as follows:
Chapter~\ref{ch:prior_work} covers related literature in greater detail than
the summary above. Chapter~\ref{ch:definitions} formalises the problem
setting and introduces the notation used in subsequent chapters.
Chapter~\ref{ch:algorithms} presents existing algorithms referenced in later
chapters as well as our algorithmic contribution, whose theoretical analysis
is presented in Chapter~\ref{ch:theoretical_results}. Our experimental
evaluation of the algorithms under different performance measures is
included in Chapter~\ref{ch:experimental_results}. Chapter
\ref{ch:discussion} concludes this thesis with a discussion of our work.

\chapter{Prior Work}
\label{ch:prior_work}
This chapter covers the most relevant algorithms for the classical
multi-armed bandit problem, as well as the prior research on the duelling
bandits problem.

\section{Multi-Armed Bandit (MAB) Problem}
\nomenclature{MAB}{Multi-Armed Bandit}
\label{sec:prior_mab}
This section summaries algorithms relevant for the remainder of this thesis
which address the finite multi-armed bandit problem. 

\subsection{Exponential-weight Algorithm for Exploration and Exploitation
(\texttt{Exp3})}
\nomenclature{\texttt{Exp3}}{Exponential-weight Algorithm for Exploration
and Exploitation}
\label{subsec:prior_exp3}
There are scenarios where the assumption of a stationary distribution is not
viable. Proposed by \cite{Auer:2003:NMB:589343.589365}, the
\textit{adversarial} or \textit{non-stationary} bandit problem makes no
assumption about the process generating the sequence of rewards. Their
simplest algorithm \texttt{Exp3} is an exponential-weights algorithm. It
maintains a vector of weights, which are mapped to a probability
distribution
\(p_t\) using the softmax function. Every round an action is sampled from
this distribution, its associated reward is observed, and the weight vector
is updated. The original algorithm is controlled through a parameter
\(\gamma \in (0, 1]\), which controls the weight of a uniform exploration
component. Other variants \citep{MAL-024} use a learning rate \(\eta\)
instead, which allows for a more elegant analysis. This work adopts the
latter approach, as seen in Algorithm~\ref{alg:exp3_unifkminus1} and
Algorithm~\ref{alg:exp3_sparring}. Both \(\gamma\) and \(\eta\) depend on
the time horizon \(T\). To make \texttt{Exp3} suitable for the anytime
setting, a variable learning rate \(\eta_t\) can be used. Assuming that the
time horizon \(T\) is known and that the rewards are bounded by the zero-one
interval, the expected regret of \texttt{Exp3} using \(\eta = \sqrt{\frac{2
\log K}{TK}}\) is bounded by \(O(\sqrt{2 K T \log K})\) \citep{MAL-024}.

\subsection{\texttt{Exp3.P}}
\label{subsec:prior_exp3p}
\texttt{Exp3} multiplies the outcome observed after playing action \(i\)
with
\(\frac 1 {p_t^i}\), compensating for the sampling process to obtain an
unbiased estimate. The lack of a lower bound of \(p_t^i\) introduces a
potentially large variance to the estimates, rendering it impossible to
derive any interesting high-probability bounds on the algorithm's regret.
\texttt{Exp3.P} \citep{Auer:2003:NMB:589343.589365} resolves this issue by
effectively adding a bias term to the aforementioned estimates. The
variation presented by \cite{MAL-024} comes along with different
high-probability bounds, depending on the parameterisation used. The
simplest bound guarantees that with probability greater than \(1 - \delta\),
the algorithm's regret is bounded by \(5.15 \sqrt{KT \log{\frac K \delta}}\)
when using the parameters required by~\cite[Theorem~3.2]{MAL-024}. This
variant of \texttt{Exp3.P} was used in Algorithm~\ref{alg:exp3p_sparring}.

\section{Duelling Bandits Problem}
\label{sec:prior_db}

As discussed in Chapter~\ref{ch:introduction} and summarised in
Table~\ref{tab:prior_work_summary}, the duelling bandits problem has been
analysed in various settings under different assumptions. As these have a
substantial impact on the nature of the problem, we structure this chapter
accordingly.

\subsection{Utility-based Setting}
\label{subsec:prior_utility}
In the utility-based setting, every action is assigned a real-valued
utility.
Outcomes of duels are modelled as Bernoulli random variables, whose bias is
determined by a known link function, which maps pairs of utilities to the
zero-one interval. This link function is assumed to induce a total order.

\subsubsection{Dueling Bandit Gradient Descent (\texttt{DBGD})}
\nomenclature{\texttt{DBGD}}{Dueling Bandit Gradient Descent}
The first formalisation of the duelling bandits problem was proposed by
\cite{Yue:2009:IOI:1553374.1553527}, who modelled the space of actions
\(\mathcal W\) as a convex, bounded, and closed space contained in a ball
with finite radius \(R\) in \(d\)-dimensional Euclidean space. The outcome
of any pairwise comparison between two actions \(w_1, w_2 \in \mathcal W\)
is modelled as an independent Bernoulli random variable with bias \(\sigma(
v(w_1) - v (w_2) )\), assuming a strictly concave utility function \(v
\colon \mathcal W \to \mathbb R\) and a rotation-symmetric, monotonic
increasing link function \(\sigma \colon \mathbb R \to [0, 1]\) with
\(\sigma(-\infty) = 0\), \(\sigma(0) = \frac 1 2\), and \(\sigma(\infty) =
1\), both functions satisfying some mild smoothness assumptions. These
assumptions give rise to a unique best action \(w^*\), which is preferred to
every other action with probability \(> \frac 1 2\). This property makes
\(w^*\) a Condorcet winner. The bias of the Bernoulli random variables
induces a gap function \(\epsilon \colon \mathcal W \times \mathcal W \to
[-\frac 1 2, \frac 1 2]\):
\begin{equation*}
  \forall w_i, w_j \in \mathcal W: \epsilon(w_i, w_j) = \sigma( v(w_i) -
  v(w_j) ) - \frac i j = \Pr \left[ w_i \succ w_j \right] - \frac i j,
\end{equation*}
which they use to define the performance measure of an algorithm selecting
pairs for comparison. The event \( w_i \succ w_j \) is equivalent to
``action \(w_i\) beats action \(w_j\) in a specific duel''. They use the
notation of regret as performance measure of an algorithm, which accumulates
the gaps between the best action in hindsight \(w^*\) and the two actions
picked. Given a finite time horizon \(T\), they define the regret after
\(T\) rounds
\begin{equation}
  R(T) = \sum_{t=1}^T \left[ \epsilon(w^*, A_t) + \epsilon(w^*, B_t) \right],
  \label{eq:def_strong_regret_yue}
\end{equation}
where \(A_t, B_t
\in \mathcal W\) denote the pair of actions picked in round \(t\).

They propose the algorithm \texttt{DBGD} and prove an upper bound on the
expected regret sublinear in \(T\) with \( O(T^{3/4} \sqrt{R d}) \). In the
information retrieval setting, DBGD allows complex parameterised retrieval
functions to be embedded in \(\mathcal W\), with the algorithm exploring
different parameterisations.

\subsubsection{Interleaved Filter (\texttt{IF})}
\nomenclature{\texttt{IF}}{Interleaved Filter}
Instead of considering a continuous space of actions, \cite{Yue:2012:KDB:2240304.2240501} 
proposed the \textit{\(K\)-armed duelling bandits problem}, a variant
considering a finite space of \(K\) actions. They coined the terms
\textit{strong regret} and \textit{weak regret}. The former corresponds to
the definition presented in~\eqref{eq:def_strong_regret_yue}, while they
defined weak regret as 
\begin{equation*}
  \widetilde R(T) = \sum_{t=1}^T \min \left\{ \epsilon(w^*, A_t),
  \epsilon(w^*, B_t) \right\}.
\end{equation*}

Adapting an explore-then-exploit approach, they proposed two tournament
elimination-based algorithms
named \texttt{IF1} and \texttt{IF2}. Both algorithms start out with a random
candidate
action \(\hat b\) and a pool of remaining actions \( W = \mathcal W
\setminus \{ \hat b \} \). They maintain an estimate \( \hat P_{\hat b, b}
\) of the candidate action's superiority to every action \(b \in W\), as
well as a confidence interval \( \hat C_{\hat b, b} \), which encompasses
the true value \( \Pr[ \hat b > b ] \) with probability greater
than \( 1 - \delta \), using \( \delta = \frac 1 {T K^2} \).
Every round, the algorithms select an action \(b\) from the pool of
remaining actions in a round-robin manner, observe the outcome of the duel
between \(\hat b\) and \(b\), and update the relevant estimate and
confidence interval. Actions are removed from \(W\) if their point estimate
of inferiority becomes greater than \( \frac 1 2 \) and \( \frac 1 2 \notin
\hat C_{\hat b, b} \).  On the other hand, if an action \(b\)'s estimate of
inferiority falls below \( \frac 1 2 \) and \( \frac 1 2 \notin \hat C_{\hat
b, b} \), then \(b\) is removed from \(W\) and replaces the candidate
action.  All estimates and confidence intervals are reset, and the
algorithms repeat until the pool \(W\) is empty, transitioning to the
exploitation phase by playing the pair \( (\hat b, \hat b) \). \texttt{IF2}
adds a pruning
step, which eliminates all actions \(b\) with \( \hat P_{\hat b, b} \succ
\frac 1 2 \) just before updating the candidate action \( \hat b \).

They proved that both the strong regret and the weak regret of \texttt{IF1}
and \texttt{IF2} are bounded
by \( O\left( \frac{K \log K}{\epsilon_{1,2}} \log T \right) \) and \(
O\left( \frac K {\epsilon_{1,2}} \log T \right)\), respectively,
\(\epsilon_{1,2} = \Pr\left[ w_1 > w_2 \right] - \frac 1 2\) denoting the
distinguishability between the two best actions \(w_1\) and \(w_2\). 

\subsubsection{\texttt{Doubler}, \texttt{MultiSBM}, \texttt{Sparring}}
\cite{Ailon:2014:RDB:3044805.3044988} introduced the notion of the
\textit{utility-based duelling bandit problem}, assuming that every action
induces a stationary
distribution with support in \([0, 1]\). A duel between two actions \(A_t,
B_t\) leads to the unobserved reward \( \frac 1 2 \left( u_t + v_t \right)
\), where \(u_t\) and \(v_t\) are sampled from the distributions induced by
the respective actions. The observable outcome is modelled as Bernoulli
random variable with bias by a linear link function
\begin{equation}
  \phi_\mathrm{lin}(u_t, v_t) = \frac{1 + v_t - u_t} 2.
  \label{eq:def_link_function_lin}
\end{equation}
They employed a utility-based definition of regret, which is related to the
strong regret \eqref{eq:def_strong_regret_yue}. This assumption allowed the
authors to provide different reductions to the classic MAB problem.

Their first algorithm, \texttt{Doubler}, is suitable for both
finite sets of \(K\) arms and infinite sets, assuming a convex utility
function similarly to \cite{Yue:2009:IOI:1553374.1553527} and a specific
link function. We will focus only on the finite case here, as this thesis
does not cover the infinite case. \texttt{Doubler} is based on exponentially
growing epochs, obsoleting the need to know a time horizon \(T\), making it
parameter-less. At its core, it uses an instance of an algorithm \(S\),
which solves a classical MAB problem. The set of actions of \(S\) is
identical to the duelling bandits problem's actions. Every round, \(S\) is
queried, yielding the first item of the pair of actions to be played. For
each epoch, \texttt{Doubler}
maintains a multi-set of actions selected by \(S\). The second action is
sampled uniformly at random from the multi-set from the previous epoch,
effectively fixing the strategy \(S\) plays against for every epoch. The
outcome of the duel between the ordered pair is observed and fed back to
\(S\) as cardinal \( \left\{ 0, 1 \right\} \) feedback, rewarding picks when
they win the duel. Assuming that the MAB algorithm employed is
\texttt{UCB1}, they showed that \texttt{Doubler} suffers at most \(
O\left(\sum_{i=2}^K \Delta_i^{-1} \log^2 T \right) \) regret in expectation,
where \(\Delta_i\) denotes the positive difference between the \(i\)-th best
action's utility and the best action's utility.

Their second contribution, \texttt{MultiSBM} (Multi Singleton Bandit
Machine\nomenclature{\texttt{SBM}}{Singleton Bandit Machine}),
improves this result by a logarithmic factor, sacrificing the ability to
handle infinite sets of actions. Instead of using a single instance of an
MAB algorithm, this approach uses \(K\) independent instances \(S_1, \dots,
S_K\), each featuring \(K\) arms. \(A_0 \in \left\{ 1, \dots, K \right\}\)
being in arbitrary action. \texttt{MultiSBM} chooses \(A_t = B_{t-1}\),
basing the first action on the second action played in the previous round.
Simultaneously, it queries \(S_{A_t}\), yielding \(B_t\). Similar to
Doubler, the outcome of the duel is fed back to \(S_{A_t}\). Using a variant
of \texttt{UCB1}, they provided an upper bound of \( O(K \log T)\) on the
expected strong regret for sufficiently large time horizons.

Lastly, the authors presented an approach
using only two instances of MAB algorithms \(S_L\) and \(S_R\), which they
named \texttt{Sparring}. Every round \(t\), \(S_L\) and \(S_R\) are queried,
yielding the actions \(A_t\) and \(B_t\), respectively. The resulting
outcome is fed back to both MAB algorithms, inverted accordingly for one of
the algorithms. While they did not provide any theoretical regret guarantees,
they included it in their experiments and claimed that it outperformed both
\texttt{Doubler} and \texttt{MultiSBM}. Due to its potential suitability for
the adversarial setting, we have included it in our empirical evaluation. 

\subsubsection{Relative \texttt{Exp3} (\texttt{REX3})}
\nomenclature{\texttt{REX3}}{Relative \texttt{Exp3}}
The idea of using two instances of an MAB algorithm was revisited by
\cite{Gajane:2015:REW:3045118.3045143}, who exmained an adversarial instance
of the utility-based duelling bandits problem with \(K\) arms. Every round,
an adversary fixes a binary utility vector \(\mathbf x(t) \in \left\{ 0, 1
\right\}^K\), which, in combination with a link function \(\phi(x_i, x_j) =
x_i - x_j \in \left\{ -1, 0, 1 \right\}\), deterministically determines the
outcome of the duel between actions \(i\) and \(j\). This ternary domain
allows the occurrence of draws. Moreover, these are guaranteed to occur if
\(K > 2\), as the assumption of the utilities being binary effectively
splits the set of actions into a subset of ``good'' actions with utility 1,
and a subset of ``bad'' actions with utility 0. Duels between actions from
the same subset result in a draw due to the assumption of the link function.
The performance measure employed is a utility-based version of strong
regret, as used by \cite{Ailon:2014:RDB:3044805.3044988}.

The algorithm they suggested, \texttt{REX3}, is based on \texttt{Sparring},
using \texttt{Exp3} as building block with some modifications. Only a single
weight vector is used, meaning that a single instance of \texttt{Exp3},
which is queried twice, plays
against itself. In case the outcome is non-zero, the weights associated with
the winner action and the loser action are increased and decreased,
respectively. If a draw is observed, no update to the weight vector is made.
Under these assumptions, they prove an upper bound on the expected strong
regret of \( 2 \sqrt{e K T \log K} = O(\sqrt{K T \log K}) \).

\subsubsection{Winner Stays (\texttt{WS})}
\nomenclature{\texttt{WS}}{Winner Stays}
Most recently, \cite{DBLP:journals/corr/ChenF16,pmlr-v70-chen17c} extended
the study of the stochastic utility-based duelling bandit problem to the
setting of weak regret. \cite{pmlr-v70-chen17c} proposed two algorithms,
Winner Stays with Weak Regret
(\texttt{WS-W}\nomenclature{\texttt{WS-W}}{Winner Stays with Weak Regret})
and Winner Stays with Strong Regret
(\texttt{WS-S}\nomenclature{\texttt{WS-S}}{Winner Stays with Strong
Regret}). \texttt{WS-W} maintains a
vector \( \mathbf C \in \mathbb Z^K \), whose elements \(C_i\) are counters
which are incremented whenever action \(i\) wins a duel, and decremented
when \(i\) loses a duel. Every round it picks \(A_t = \arg \max_i C_i\),
breaking ties by preferring \(A_{t-1}\) and \(B_{t-1}\). In case none
of the previous actions played maximise \(C_i\), \(A_t\) is sampled
uniformly from \( \arg \max_i C_i \). \(B_t\) is selected in a similar
manner, excluding \(A_t\) from the actions available. Upon observation of
the outcome the relevant counters are updated.

\texttt{WS-S} builds on this algorithm
and makes it applicable to the strong regret setting. Splitting the time
axis in exponentially growing epochs, it uses \texttt{WS-W} for the first
part of
every epoch. This fraction is controlled by a parameter \(\beta\) and
depends on the epoch. The second part of the epoch is an exploitation phase,
which uses the best action determined by the precedent exploration phase. 
While they did not reference utilities directly, their theoretical analysis
of the expected regret proves tighter bounds when assuming a total order,
namely \(O(K \log K)\) for weak regret and \(O(K \log (KT)\) for strong
regret.

\subsubsection{Comparing The Best (\texttt{CTB})}
\nomenclature{\texttt{CTB}}{Comparing The Best}
\cite{DBLP:journals/corr/ChenF16} made the assumption that each arm \(i\)
has an observable \(d\)-dimensional feature vector \(\mathbf A_i\). Assuming
an unknown vector \(\bm \theta \in \mathbb R^d\), the known utility
function \( u(\bm \theta, \mathbf A_i) = \left< \bm \theta, \mathbf
A_i \right> \) assigns real-valued utilities to them. An exploitable
dependence between the arms as well as a total order is induced by a link
function. The authors suggest both the Bradley-Terry model
\begin{equation}
  \Pr\left[ i \succ j \right] = \frac{\exp( u(\bm \theta, \mathbf A_i)
  )}{\exp( u(\bm \theta, \mathbf A_i) + u(\bm \theta, \mathbf A_j) )}
  \label{eq:def_link_function_BT}
\end{equation}
and the probit model
\begin{equation}
  \Pr\left[ i \succ j \right]
  = \Phi(u(\bm \theta, \mathbf A_i) - u(\bm \theta, \mathbf A_j)),
  \label{eq:def_link_function_probit}
\end{equation}
which uses the cumulative distribution function of the normal distribution.

Their algorithm Comparing The Best assigns a score to all \(K!\) possible
orders, which relates to
the posterior distribution of an order being the relevant order based
on all previous observations. Every round \texttt{CTB} picks the
respective best
action of the two orders with maximal score, observes the outcome of the
duel, and updates all scores. The authors provided an upper bound of \(
O(K^2) \) on the expected weak regret, ignoring any instance-dependent
multiplicative constants. Using synthetic datasets, \texttt{CTB}
outperformed WS-W as
well as and other algorithms which were not explicitly designed for the weak
regret setting.

\subsection{Condorcet Winner}
The Condorcet winner setting only makes the assumption that the best arm
wins against all other arms, dropping the requirement of a total order.
It is therefore a generalisation of the utility-based setting, increasing
the model's applicability.

\subsubsection{Beat-The-Mean (\texttt{BTM})}
\nomenclature{\texttt{BTM}}{Beat-The-Mean}
\cite{Yue:2011:BMB:3104482.3104513} were the first to propose and examine
this relaxed setting. They invented the algorithm Beat-The-Mean, which does
not require any
transitivity assumptions. A working set of arms \(W\) contains initially all
\(K\) arms. Over \(K - 1\) epochs, the action in \(W\) which has been
involved in the least number of comparisons is picked to be \(A_t\). \(B_t\)
is sampled uniformly at random from the working set. Keeping track of the
fraction of duels won and duels played, the action with the lowest empirical
performance is removed from the working set when it is ruled out as the
Condorcet winner with sufficient confidence. After adjusting the counters
keeping track of the individual arms' performance, the next epoch begins. In
case just a single action is left, \texttt{BTM} transitions to its
exploitation phase. Ignoring any instance-dependent multiplicative
constants, its strong regret in expectation is bounded by \( O(K \log T) \).
This hides a multiplicative factor of \(\frac{\gamma^7}{\epsilon_*}\), with
\(\gamma \geq 1\) increasing when transitivity of the actions' preferences
is violated, and \(\epsilon_* = \min_i \epsilon(w^*, i)\). 

\subsubsection{Sensitivity Analysis of VAriables for Generic Exploration
(\texttt{SAVAGE})}
\nomenclature{\texttt{SAVAGE}}{Sensitivity Analysis of VAriables for Generic
Exploration}
Unlike all prior work covered so far in this chapter,
\cite{Urvoy:2013:GEK:3042817.3042904} examines the stochastic duelling
bandits problem in a modified version of the PAC setting. Instead of
bounding some notion of regret, they provide bounds on the number of rounds
\texttt{SAVAGE} requires to terminate, yielding the \(\varepsilon\)-correct
arm with probability greater than \(1 - \delta\).

Instead of making any explicit assumptions about the winner, \texttt{SAVAGE}
uses a
``sensitivity analysis subroutine'', which incorporates the definition of
the winner, and is used to gradually reduce the working set until all other
actions have been eliminated. In the Condorcet winner setting, the
winner is found with probability greater than \(1 - \delta\) in \( O\left(
K^2 \log\left( \frac{KT} \delta \right) \right) \) when loosening the bound
to adjust for instance-dependent constants.

\subsubsection{Relative Upper Confidence Bound (\texttt{RUCB})}
\nomenclature{\texttt{RUCB}}{Relative Upper Confidence Bound}
Similarly to \texttt{UCB}\nomenclature{\texttt{UCB}}{Upper Confidence Bound}
\citep{Auer:2002:FAM:599614.599677}, \texttt{RUCB} uses a time-dependent
upper confidence bound on the
actions' performance, yielding regret guarantees which hold for any time
\(t\). The algorithm maintains a matrix \( \mathbf W \in \mathbb Z_+^{K
\times K} \), whose elements \( W_{ij} \) keep track of the number of rounds
action \(i\) won against action \(j\). Given a fixed input parameter
\(\alpha > \frac 1 2\), it constructs a corresponding upper confidence
matrix \(\mathbf U \in \mathbb R_+^{K \times K}\) with
\begin{equation}
  U_{ij} = \frac{W_{ij}}{W_{ij} + W_{ji}} + \sqrt{\frac{\alpha \log
  t}{W_{ij} + W_{ji}}}
  \label{eq:def_rucb_ucb}
\end{equation}
if \(i \neq j\) and \(U_{ii} = \frac 1 2\). The first term of
\eqref{eq:def_rucb_ucb} rewards well-performing actions, while the second
term boosts less-played actions, guaranteeing sufficient exploration. A set
\( \mathcal C = \left\{ i \middle| \forall j: U_{ij} \geq \frac 1 2 \right\}
\) is updated every round. If it is empty, \(A_t\) is picked uniformly at
random from the set of actions \(\left\{ 1, \dots, K \right\}\). If it
contains just a single element, \(A_t\) assumes this action. Otherwise it is
sampled from \(\mathcal C\) using a custom distribution. Using \(B_t = \arg
\max_j U_{j A_t} \), the outcome of the duel between \(A_t\) and \(B_t\) is
observed and \(\mathbf W\) is updated.

The authors provided an upper bound on the expected strong regret of \( O( K
\log t) \), which is consistent with the lower bound of the duelling bandits
problem in this setting \citep{Yue:2012:KDB:2240304.2240501}.

\subsubsection{Relative Minimum Empirical Divergence (\texttt{RMED})}
\nomenclature{\texttt{RMED}}{Relative Minimum Empirical Divergence}
\cite{pmlr-v40-Komiyama15} provided a range of algorithms which explicitly
make use of the binary Kullback-Leibler divergence. After an initial
exploration of all pairs of actions, \texttt{RMED} proceeds to an
exploration / exploitation phase. Due to its complexity, we refer to the
original paper at this point. The simplest algorithm, \texttt{RMED1}, comes
along with a proof of an upper bound on its expected strong regret of \(O(K
\log T)\). The
more complicated algorithm \texttt{RMED2FH} comes along with a proof sketch,
which promises a regret bound matching the lower bound provided in the same
paper, improving the previously proved bound by a constant factor.

\subsubsection{Winner Stays (\texttt{WS})}
When dropping the assumption of a total order, as described in
Subsection~\ref{subsec:prior_utility}, \cite{pmlr-v70-chen17c} showed
that the expected weak regret of \texttt{WS-W} is bounded by \(O(K^2)\),
while the expected strong regret of \texttt{WS-S} has an upper bound of
\(O(K^2 + K \log T)\).

\subsection{Borda Winner}
Unlike the Copeland winner and the von-Neumann winner, which pose a
generalisation of the Condorcet winner, the Borda winner is the action
maximising the probability of winning a duel sampled from all \(K\) actions
uniformly at random. This dependence on all other actions implies that the
Borda winner is not guaranteed to be the Condorcet winner when the
assumption of a total order is not fulfilled.

\subsubsection{Sensitivity Analysis of VAriables for Generic Exploration
(\texttt{SAVAGE})}
While the \cite{Urvoy:2013:GEK:3042817.3042904} were the first to suggest
the Borda winner and pointed out that \texttt{SAVAGE} was directly
applicable in this setting, they included the Borda winner only in their
experimental section.  The authors argued that this setting was less complex
than the Copeland setting, and could be readily solved by classical MAB
algorithms.

\subsubsection{Successive Elimination with Comparison Sparsity
(\texttt{SECS})} \nomenclature{\texttt{SECS}}{Successive Elimination with
Comparison Sparsity}
\cite{pmlr-v38-jamieson15} used the suggestion by
\cite{Urvoy:2013:GEK:3042817.3042904} to employ a classical MAB algorithm to
solve the Borda duelling bandits problem as a baseline, which they termed
\textit{Borda reduction}. \texttt{SECS} exploits different structural
assumptions of the constant preference matrix, allowing the identification
of the Borda winner in fewer rounds than the Borda reduction, if fulfilled.

\subsection{Copeland Winner}
Instead of requiring an action winning against all other actions with
probability greater than \(\frac 1 2\), the Copeland winner is any action
winning against the greatest number of other actions, not taking into
account by how much it outperforms them.

\subsubsection{Sensitivity Analysis of VAriables for Generic Exploration
(\texttt{SAVAGE})}

When applying \texttt{SAVAGE} under the assumption of the existence of a
Copeland winner, \cite{Urvoy:2013:GEK:3042817.3042904} bound the number of
rounds the algorithm requires to yield the winner with probability greater
than \(1 - \delta\) by \( O \left( K^2 \log \left( \frac{KT} \delta \right)
\right) \).

\subsubsection{Copeland Confidence Bounds (\texttt{CCB}), Scalable Copeland
Bandits (\texttt{SCB})}
\nomenclature{\texttt{CCB}}{Copeland Confidence Bounds}
\nomenclature{\texttt{SCB}}{Scalable Copeland Bandits}
\cite{Zoghi:2015:CDB:2969239.2969274} improved on the result above using
\texttt{CCB}, which is inspired by RUCB. In addition to the upper confidence
bound
matrix \(\mathbf U\), it introduces a lower confidence bound matrix
\(\mathbf L\) and a counter \(L_C \in \mathbb N\) estimating the number of
arms the best arm loses against. Furthermore, it maintains a working set
\(\mathcal B_1\) of potential Copeland winners and, for every arm \(i\), a
set of arms \(\mathcal B_1^i\) potentially beating \(i\). Every
round, \texttt{CCB} chooses \(A_t\) to be an empirically well-performing
arm, with arms in \(\mathcal B_1\) more likely to be selected. \(B_t\) is
chosen from
\(\mathcal B_1^{A_t}\) with the intention of ultimately removing either
\(A_t\) from \(\mathcal B_1\) (if \(A_t \in \mathcal B_1\)) or \(B_t\) from
\(\mathcal B_1^{A_t}\). The algorithm has a multitude of mechanisms
controlling the specific mechanism according to which these actions are
sampled, which are out of the scope of this summary.

Assuming that there exists no pair of distinct actions \(i, j\) for which
\(\Pr\left[ i \succ j \right] = \frac 1 2\), the expected strong regret of
\texttt{CCB} is bounded by \( O(K^2 + K \log T) \). To remove the quadratic
dependence on \(K\), which limits the applicability for problem instances
with a large number of arms, \texttt{SCB} divides the time axis into epochs
of iterated exponentially length. Every epoch, it launches an exploration
phase using a variant of \texttt{CCB}. Upon termination of this subroutine,
it proceeds with the exploitation phase for the remainder of the epoch,
playing twice the approximated Copeland winner. \texttt{SCB} comes along
with a upper bound on the expected strong regret of \( O(K \log K \log T)
\), suggesting better results for larger values of \(K\).

\subsubsection{Double Thompson Sampling (\texttt{DT-S})}
\nomenclature{\texttt{DT-S}}{Double Thompson Sampling}
Motivated by previous applications of Thompson sampling to the MAB problem
\citep{Chapelle:2011:EET:2986459.2986710}, \cite{NIPS2016_6157} presented
\texttt{DT-S}, which uses Thompson sampling to select both actions. They
proved an upper bound of \( O(K^2 \log T) \) on the expected strong regret
in the Copeland setting.

\subsection{Von-Neumann Winner}
The von-Neumann winner stems from a game-theoretic interpretation of the
duelling bandits problem. Instead of maximising a pre-defined quality
measure such as the probability of winning against an action sampled
uniformly at random or the number of actions an arm wins against with
probability \(> \frac 1 2\), the von-Neumann winner is the distribution over
all arms winning against any other policy with probability \(> \frac 1 2\)
\citep{pmlr-v40-Dudik15}. This definition has two convenient properties.
First, it is compatible to the definition of the Condorcet winner. Secondly,
unlike the Borda winner and the Copeland winner, it is not influenced by the
existence of clones of actions \citep{pmlr-v40-Dudik15}.

\subsubsection{\texttt{Sparring Exp4.P}, \texttt{SparringFPL},
\texttt{ProjectedGD}}
In addition to examining the duelling bandits problem in a novel winner
scenario, \cite{pmlr-v40-Dudik15} suggested a variant which incorporates
context, similar to the classical MAB with expert advice
\citep{Auer:2003:NMB:589343.589365}. They proposed an algorithm of
\texttt{Sparring} with two independent instances of \texttt{Exp4.P}, which
is a variant of \texttt{Exp3.P} capable of incorporating context
\citep{MAL-024}. The authors argued that the strong regret of
\texttt{Sparring Exp4.P} is bounded by \( O\left( \sqrt{K T \frac{\lvert \Pi
\rvert} \delta }\right) \) with probability greater than \( 1 - \delta \),
where \( \lvert \Pi \rvert \) denotes the size of the policy space \( \Pi
\). This bound is holds also for the adversarial setting.

\texttt{SparringFPL} and \texttt{ProjectedGD} are designed for large policy
spaces, offering better time and space requirements. We omit them for this
summary as this work's focus does not lie on the contextual problems.

\subsubsection{Sparse Sparring (\texttt{SPAR2})}
\nomenclature{\texttt{SPAR2}}{Sparse Sparring}

\cite{pmlr-v49-balsubramani16} explored the stochastic duelling bandits
problem in the context-less setting. Based on the assumption that the
von-Neumann winner has only support by small number of actions \(s \ll K\),
they suggested \texttt{SPAR2}, which employs a \texttt{Sparring} algorithm
using two instances of \texttt{Exp3.P}. Similar to \texttt{CCB}, it uses
both an upper and a lower confidence bound on the frequentist estimates of
the elements of the preference matrix. These are used to eliminate actions
which are likely not to be included in the von-Neumann winner's support,
which are in turn removed from the \texttt{Sparring} instance. Ignoring any
instance-dependent additive constants, \cite{pmlr-v49-balsubramani16}
provide bounds on the strong regret of \texttt{SPAR2} of \( \tilde
O(\sqrt{sT}) \).

\chapter{Definitions}
\label{ch:definitions}
This chapter lays the foundation for the remaining chapters by providing a
unified framework for classifying a range of variations of the duelling
bandits problem. The learner is presented with a fixed set of \(K > 1\)
actions and a time horizon \(T \in \mathbb N \cup \left\{ \infty \right\}\).
The environment fixes a sequence of skew-symmetric outcome matrices \(
  \mathcal M = \left( \mathbf M^t \right)_{t=1}^T \) with 
\begin{equation}
  \mathbf M^t = -\left( \mathbf M^t \right)^T \in \left\{ -1, 0, 1
\right\}^{K \times K}
  \label{eq:def_Mt}
\end{equation}
such that
\begin{equation*}
  M_{ij}^t = \begin{cases}
    -1  & \text{if action \(i\) loses against action \(j\)} \\
    0   & \text{if \(i = j\)} \\
    1   & \text{if action \(i\) wins against action \(j\)}
  \end{cases}.
\end{equation*}
Some problem formulations allow ties between non-equal actions, e.g.
\cite{Gajane:2015:REW:3045118.3045143}. We explicitly disallow this
behaviour. In addition to the sequence of outcome
matrices, the environment fixes a sequence of loss vectors:
\begin{equation*}
  \mathcal L = \left( \bm \ell_t \right)_{t=1}^T \text{ with } \bm \ell_t
  \in [0, 1]^K.
\end{equation*}
We will index the loss vectors as \( \ell_t^a\) with \( a \in \left\{ 1,
\hdots, K \right\}\). Neither of these sequences is revealed to the learner.
Instead, the learner follows the following protocol for every round \(t = 1,
\hdots, T\):
\begin{enumerate}
  \item Pick \(A_t, B_t \in \left\{ 1, \hdots, K \right\}\), inducing a pair
    of losses \( \ell_t^{A_t}, \ell_t^{B_t} \).
  \item Observe outcome \( Y_t = M_{A_t B_t}^t \).
\end{enumerate}
The learner's objective is to minimise some notion of regret. Assuming a
finite time horizon \(T\), the finite-time regret can be formulated as
\begin{equation}
  R(T) = \sum_{t=1}^T \psi\left(\ell_t^{A_t}, \ell_t^{B_t}\right) - \min_a
  \sum_{t=1}^T \ell_t^a
  \label{eq:def_regret_T}
\end{equation}
with \(\psi \colon [0,1] \times [0,1] \to [0,1]\). This general framework
allows the classification of settings of the duelling bandits problem by
three independent parameters:
\begin{enumerate}
  \item The process underlying the generation of sequence of outcome
    matrices (Section~\ref{sec:def_outcome_generation}).
  \item The process underlying the generation of the sequence of loss
    vectors (Section~\ref{sec:def_loss_generation}).
  \item The definition of regret, described by \(\psi\)
    (Section~\ref{sec:def_regrets}).
\end{enumerate}

\section{Generation of Outcomes}
\label{sec:def_outcome_generation}
The process generating the sequence of outcome matrices is characterised by
the following two properties:
\begin{enumerate}
  \item If the individual outcomes are sampled from a stationary
    distribution, the duelling bandits problem is said to take place in the
    stochastic setting. If the distribution is non-stationary, which covers
    the case of the outcomes being selected in an arbitrary manner, the
    setting is referred to as adversarial or non-stochastic.
  \item Most settings discussed in the literature base the outcome matrix
    generation process on a utility vector \( \mathbf x_t \in \mathbb R^K
    \), which, in combination with a link function \( \phi \colon \mathbb R
    \times \mathbb R \to [0, 1] \), models the individual outcomes as a
    Bernoulli random variable with bias induced by the link function known
    to the learner:
    \begin{equation*}
      M_{ij}^t = -M_{ji}^t = 2 \widetilde M_{ij}^t - 1 \text{ with }
      \widetilde M_{ij}^t \sim B \left( 1, \phi \left( x_t^i, x_t^j \right)
      \right).
    \end{equation*}
    This assumption gives rise to the setting known as the utility-based
    duelling bandits problem, which induces a total order of the arms for a
    given round. On the other hand, if no such limitations are present, we
    denote the setting as non-utility-based.
\end{enumerate}

When designing algorithms for the utility-based setting, the link function
is assumed to be known. This setup is compatible to the Bradley-Terry model
\eqref{eq:def_link_function_BT}, the probit model
\eqref{eq:def_link_function_probit}, and the linear model
\eqref{eq:def_link_function_lin}. We will restrict ourselves to the latter.
Like \cite{Ailon:2014:RDB:3044805.3044988}, we will bound the utilities by
the zero-one interval and use the linear link function
\begin{equation}
  \phi(x_i, x_j) = \frac{1 + x_i - x_j} 2.
  \label{eq:def_link_function_linear}
\end{equation}
This causes the outcome \( Y_t \) to be an unbiased estimator of the
difference of utilities with respect to the randomness of the sampling
process of the Bernoulli random variable \( \widetilde M_{ij}^t \):
\begin{equation*}
  \mathbb E \left[ Y_t \middle| A_t, B_t \right] =
  \frac{1 + x_t^{A_t} - x_t^{B_t}} 2 - \frac{1 + x_t^{B_t} - x_t^{A_t}} 2
  = x_t^{A_t} - x_t^{B_t}.
  \label{eq:bt_pr}
\end{equation*}

\section{Generation of Loss Vectors}
\label{sec:def_loss_generation}
Similar to the classical multi-armed bandit problem \citep{MAL-024}, we will
use the following relation between rewards and losses for the utility-based
setting, treating rewards and utilities synonymously:
\begin{equation*}
  \bm \ell = \mathbf 1 - \mathbf x.
  \label{eq:def_utility_loss}
\end{equation*}
This allows the interpretation that in the utility-based setting the
environment selects the loss vectors, which in turn are used to generate the
outcome matrix using a known stochastic process.

The best action in hindsight is the one minimising the cumulative
utility-based loss\footnote{We use subscripts to denote different types of
winners.}:
\begin{align}
  a_U^* = \arg \min_a \sum_{t=1}^T \ell_t^a.
  \label{eq:def_best_utility}
\end{align}

For the non-utility-based setting the definition of loss is less
straight-forward, as there are multiple, conflicting winner criteria.  The
remainder of this section covers multiple ways of translating a sequence of
outcome matrices \( \left\{ -1, 0, 1 \right\}^{K \times K \times T}\) and a
round index \(t\) to a loss vector \(\bm \ell \in [0, 1]^K\).

\subsection{Borda Winner}
The Borda setting is a natural generalisation of both the stochastic
duelling bandit problem and the non-stochastic utility-based duelling bandit
problem. We will focus on the loss induced by the normalised Borda count,
which is the probability of arm \(i\) beating a second arm sampled uniformly
at random \citep{Urvoy:2013:GEK:3042817.3042904}:
\begin{equation*}
  U_{Bor}(i) = \frac 1 K \sum_{j = 1}^K p_{ij}^t,
\end{equation*}
with \(p_{ij}^t\) denoting the probability of arm \(i\) beating arm \(j\) in
a specific round \(t\). In the non-stochastic setting, this leads to the
following definition of loss vectors \(\bm \ell_t\):
\begin{equation}
  \ell_t^i = \frac 1 K \sum_{j=1}^K \frac{M_{ji}^t + 1}{2}
  = \frac 1 2 + \frac 1 {2 K} \sum_{j=1}^K M_{ji}^t.
\label{eq:def_loss_borda}
\end{equation}
The Borda loss depends at any round \(t\) depends only on the outcome matrix
\(\mathbf M^t\). The Borda winner is an action minimising the cumulative
Borda loss:
\begin{equation*}
  a_B^* = \arg \min_a \sum_{t=1}^T \ell_t^a.
\end{equation*}

\subsection{Copeland Winner}
\label{subsec:def_copeland}
The Copeland winner is the action winning against the largest number of
other actions. \cite{Zoghi:2015:CDB:2969239.2969274} assumed that the
outcomes are sampled from a stationary distribution for every pair of
actions and defined the Copeland score for an arm \(i\) as the number of
other arms it wins against with probability \(> 0.5\):
\begin{equation*}
  U_{Cpld}(i) = \sum_{j \in [K] \setminus \left\{ i \right\}} \mathbbm
  1(p_{ij} > 0.5)
\end{equation*}
with \(p_{ij}\) denoting the probability of arm \(i\) winning against arm
\(j\). They also introduce the normalised Copeland score \(U_{cpld}(i)
\in [0,1]\):
\begin{equation*}
  U_{cpld}(i) = \frac {U_{Cpld}} {K - 1} = \frac 1 {K - 1} \sum_{j \in [K]
  \setminus \left\{ i \right\}} \mathbbm 1(p_{ij} > 0.5).
\end{equation*}
The following adaptations are based on the normalised Copeland score. Unlike
the Borda loss, which was based on the instantaneous outcome matrix, we base
the Copeland score on the cumulative outcome matrix \( \mathbf M(T) =
\sum_{t=1}^T \mathbf M^t \). This renders the loss vector \(\ell_t\)
constant for all \(t\) with
\begin{equation}
  \ell_t^i = \frac 1 {K - 1} \sum_{j=1}^K \mathbbm 1 \left( \left[ \mathbf
  M(T)\right]_{ij} < 0 \right).
  \label{eq:def_loss_copeland}
\end{equation}
A Copeland winner is an action minimising the cumulative loss:
\begin{equation*}
  a_C^* = \arg \min_a \sum_{t=1}^T \ell_t^a.
\end{equation*}
The independence of \(t\) means that the loss vector depends only on the
cumulative outcome matrix. If one were to base the Copeland loss on the
individual outcome matrices instead, i.e.
\begin{equation*}
  \ell_t^i = \frac 1 {K-1} \sum_{j=1}^K \mathbbm 1 \left( M_{ij}^t < 0
  \right),
\end{equation*}
due to skew-symmetry of the outcome matrices \eqref{eq:def_Mt}, the loss
definition can be rewritten as a linear transformation of the Borda loss
\eqref{eq:def_loss_borda}:
\begin{align*}
  \ell_t^i &= \frac 1 {K-1} \sum_{j=1}^K \mathbbm 1\left( M_{ij}^t < 0
  \right) \\
  &= \frac 1 {2 (K - 1)} \left( K - 1 - \sum_{j=1}^K M_{ij} \right) \\
  &= \frac 1 2 - \frac 1 {2(K - 1)} \sum_{j=1}^K M_{ij} \\
  &= \frac 1 2 + \frac 1 {2(K - 1)} \sum_{j=1}^K M_{ji},
\end{align*}
with the second equality following from
\begin{align*}
  K - 1 &= \sum_{j=1}^K \mathbbm 1 \left( M_{ij}^t > 0 \right)
  + \sum_{j=1}^K \mathbbm 1 \left( M_{ij}^t < 0 \right) \\
  \sum_{j=1}^K M_{ij}^t &= \sum_{j=1}^K \mathbbm 1 \left( M_{ij}^t > 0
  \right) - \sum_{j=1}^K \mathbbm 1 \left( M_{ij}^t < 0 \right) \\
  &= K - 1 - 2 \sum_{j=1}^K \mathbbm 1 \left( M_{ij}^t < 0 \right) \\
  \implies \sum_{j=1}^K \mathbbm 1 \left( M_{ij}^t <0 \right)
  &= \frac 1 2 \left( K - 1 - \sum_{j=1}^K M_{ij}^t \right).
\end{align*}
This implies that the winner induced by this
definition of the Copeland loss is identical to the Borda winner, and the
losses can be converted by a linear transformation. Definition
\eqref{eq:def_loss_copeland} does not cause any of these problem, justifying
our choice to base the Copeland loss only on the cumulative outcome matrix.
On the other hand, basing the Copeland loss on the cumulative outcome matrix
makes it sensitive to small changes in case the accumulated outcome \(
\sum_{t=1}^T M_{ij} \) is close to zero, as demonstrated by the following
sketch.
Given a sequence of outcome matrices \( \left( \mathbf M^t \right)_{t=1}^{T
- 1} \) with actions 1 and 2 making up the set of Copeland winners and
\(\sum_{t=1}^{T - 1} M_{1,2}^t = 0\), a sequence \( \left( \mathbf M^t
\right)_{t=1}^T \) which chooses \( M_{1,2}^T \sim B(1, 0.5) \) can induce
expected regret linear in \(T\), as playing the non-Copeland winner induces
instantaneous regret \( > \frac 1 {K - 1} \). As any algorithm \( \mathcal A
\) picks the right action with probability \( \leq 0.5 \), \( \mathcal A \)
suffers at least \( O\left(\frac T {2 (K - 1)}\right) = O(T) \) regret for
some sequences. This renders the duelling bandits problem using our
definition of Copeland regret infeasible.

These considerations are superfluous when considering the Borda loss, as the
resulting definition is identical to our definition
\eqref{eq:def_loss_borda} due to associativity and commutativity of
additivity.

\subsection{Von-Neumann Winner}
\cite{pmlr-v40-Dudik15} have introduced the notion of the von-Neumann winner to
overcome the aforementioned limitations of the Borda winner and the Copeland
winner. 
We define the von-Neumann winner as the stationary distribution over actions
\(\mathbf u \in \Delta^K\), minimising the opponent's payoff in hindsight:
\begin{equation*}
  \min_{\mathbf u} \max_{\mathbf v \in \Delta^K} \mathbf u^T \mathbf M(T)
  \mathbf v
  = \min_{\mathbf u} \max_i \left[ \mathbf u^T \mathbf M(T) \right]_i
\end{equation*}
The reward in this setting can be modelled as the sum of expected rewards
obtained when playing against the von Neumann winner. Playing against the
von Neumann winner leads to the following loss in expectation:
\begin{equation*}
  \ell_t^i = \left[ \mathbf u^T \mathbf M^t \right]_i
\end{equation*} 
The loss sequence of a sequence of
actions reflects the number of duels won against the von-Neumann \(\mathbf
u\) winner in expectation. This loss sequence depends on the von-Neumann
winner, and as the problem can be modelled as a zero-sum game, we have
\(\mathbb E \left[ \sum_{t=1}^T \ell_t^{I_t \sim \mathbf u} \right] = 0\).
This simplifies the definition of regret \eqref{eq:def_regret_T} in the
von-Neumann setting:
the definitions for both the strong and the weak regret:
\begin{equation}
  R(T) = \sum_{t=1}^T \psi \left( \ell_t^{A_t}, \ell_t^{B_t} \right).
  \label{eq:def_regret_von_neumann}
\end{equation}

\section{Notions of Regret}
\label{sec:def_regrets}
As described in \eqref{eq:def_regret_T}, the notion of regret relates the
sequence of pairs of losses to the minimal cumulative loss induced by a
equation action. Throughout the literature, two measures of regrets have
been discussed:
\begin{enumerate}
  \item Based on the definition of regret initially suggested by
    \cite{Yue:2009:IOI:1553374.1553527}, the most common notion of strong regret is based on
    the mean of the pairwise losses \footnote{Both
      \cite{Yue:2009:IOI:1553374.1553527} and
      \cite{Yue:2012:KDB:2240304.2240501} differs from this definition by a factor of two. As
      their analyses ignore any multiplicative constants, we decided to use
      the more common definition of strong regret, as introduced by
    \cite{Yue:2011:BMB:3104482.3104513}.}
    \begin{equation*}
      \psi_S(x, y) = \frac 1 2 \left( x + y \right),
    \end{equation*}
    yielding the following definition of finite-time strong regret:
    \begin{equation*}
      R_S(T) = \frac 1 2 \sum_{t=1}^T \left( \ell_t^{A_t} + \ell_t^{B_t} \right) -
      \min_a \sum_{t=1}^T \ell_t^a.
    \end{equation*}
  \item Further work by \cite{Yue:2012:KDB:2240304.2240501} suggested a novel definition of
    regret, denoted as weak regret. Using
    \begin{equation}
      \psi_W(x, y) = \min \left\{ x, y \right\},
      \label{eq:def_psi_weak}
    \end{equation}
    the accompanying definition of regret becomes
    \begin{equation}
      R_W(T) = \sum_{t=1}^T \min \left\{ \ell_t^{A_t}, \ell_t^{B_t} \right\} -
      \min_a \sum_{t=1}^T \ell_t^a.
      \label{eq:def_regret_weak}
    \end{equation}
\end{enumerate}

\chapter{Algorithms}
\label{ch:algorithms}
This chapter describes our algorithm as well as adaptions of other
implementations we used in experiments in
Chapter~\ref{ch:experimental_results}.

\section[\nameref{alg:exp3_unifkminus1}]{\nameref{alg:exp3_unifkminus1}
(Main Contribution)}
\label{sec:alg_exp3_unifkminus1}
The duelling bandits problem with strong regret forces the learner to
approximate the winner using both arms to get a meaningful bound. The weak
regret relaxes this requirement, which can be leveraged in different ways.
As only a single element of the pair of actions played is required to
imitate the winner, the other action can be used for exploration. A similar
problem has been suggested by \cite{1807.00636}, who examined a variation of
the classical multi-armed bandit problem. Every round, after observing and
suffering the loss of an action \(A_t\), the modified game protocol allows
the learner to observe the loss of a second action without suffering its
loss. This problem statement differs from the duelling bandits problem in
two aspects. On the one hand, the weak regret in our setting is agnostic to
the order, making it slightly more tolerant. On the other hand, we observe
only binary feedback, while \cite{1807.00636} assumed a pair of real values.
This is a significant difference, as the authors' primary motivation  was to
leverage the boundedness of the effective range of losses, allowing for a
better instance-dependent bound on the regret. While this prohibits the
practicability of any of their work's details, we use a similar approach to
their \textit{Second Order Difference Adjustments}
(\texttt{SODA}\nomenclature{\texttt{SODA}}{Second Order Difference
Adjustments}) algorithm. Using an \texttt{Exp3}-based algorithm to determine
the first action \(A_t\), \nameref{alg:exp3_unifkminus1} samples the second
action \(B_t\) uniformly from the remaining actions.
Algorithm~\ref{alg:exp3_unifkminus1} describes the implementation in detail.
\\
\begin{algorithm}[H]
  \algotitle{\texttt{Exp3+UnifK-1}}{alg:exp3_unifkminus1.title}
  \TitleOfAlgo{\texttt{Exp3+UnifK-1}}
  \KwIn{ Learning rate \(\eta\)}
  \(\forall a: \widetilde L_0(a) = 0\) \\
  \For{\(t = 1, 2, \hdots\)}{
    \(\forall a: p_t(a) = \frac{\exp \left( - \eta \widetilde L_{t-1}(a)
      \right)}
    {\sum_{i=1}^K \exp \left( - \eta \widetilde L_{t-1}(i) \right)}\) \\
    Sample \(A_t \sim p_t\) \\
    Sample \(B_t\) uniformly from remaining elements \([K] \setminus \left\{
    A_t \right\}\) \\
    Play \( (A_t, B_t) \) and observe \(Y_t\) \\
    \(\forall a: \widetilde L_t(a) = \frac{\mathbbm 1(a = A_t)}{2p_t^{A_t}}
    (1 - Y_t)\)
  }
  \caption{\texttt{Exp3+UnifKminus1}}
  \label{alg:exp3_unifkminus1}
\end{algorithm}

\section[\nameref{alg:exp3_sparring}]
{\nameref{alg:exp3_sparring} \citep{Ailon:2014:RDB:3044805.3044988}}
\label{sec:alg_exp3_sparring}
\texttt{Sparring} was suggested as a heuristic generic algorithm
\citep{Ailon:2014:RDB:3044805.3044988}. We include
Algorithm~\ref{alg:exp3_sparring} in our experimental section using two
instances of \texttt{Exp3} based on the implementation and parameterisation
by \cite{MAL-024}. We conjecture that \nameref{alg:exp3_sparring}
approximates the von-Neumann winner, as argued for \texttt{Sparring Exp4.P}
\citep{pmlr-v40-Dudik15}.
\\
\begin{algorithm}[H]
  \algotitle{\texttt{Exp3-Sparring}}{alg:exp3_sparring.title}
  \TitleOfAlgo{\texttt{Exp3-Sparring}}
  \KwIn{ Learning rate \(\eta_t\)}
  \(\forall a: \widetilde L_0^A(a) = L_0^B(a) = 0\) \\
  \For{\(t = 1, 2, \hdots\)}{
    \(\forall a: p_t^A(a) = \frac{\exp \left( - \eta_t \widetilde
      L_{t-1}^A(a) \right)} {\sum_{i=1}^K \exp \left( - \eta_t \widetilde
      L_{t-1}^A(i) \right)}, p_t^B(a) = \frac{\exp \left( - \eta_t \widetilde
      L_{t-1}^B(a) \right)} {\sum_{i=1}^K \exp \left( - \eta_t \widetilde
    L_{t-1}^B(i) \right)}\)
    \\
    Sample \(A_t \sim p_t^A, B_t \sim p_t^B\) \\
    Play \( (A_t, B_t) \) and observe \(Y_t\) \\
    \(\forall a: \widetilde L_t^A(a) = \frac{\mathbbm 1(a = A_t)}{2p_t^{A_t}}
    (1 - Y_t), \widetilde L_t^B(a) = \frac{\mathbbm 1(a = B_t)}{2p_t^{B_t}}
    (1 + Y_t) \)
  }
  \caption[\texttt{Exp3-Sparring}]{\texttt{Exp3-Sparring}, as suggested by
  \cite{Ailon:2014:RDB:3044805.3044988} using \texttt{Exp3} implementation
and parameterisation by \cite{MAL-024}}
  \label{alg:exp3_sparring}
\end{algorithm}

\section{\nameref{alg:exp3p_sparring}}
\label{sec:alg_exp3p_sparring}

We are unaware of any work which has derived any explicit guarantees for
\nameref{alg:exp3_sparring}. However, \cite{pmlr-v40-Dudik15} presented
asymptotic bounds their context-incorporating algorithm \texttt{Sparring
Exp4.P}. Assuming \(K\) experts with unequal stationary distributions, each
having support of only a single actions, we effectively remove the expert
advice, yielding Algorithm~\ref{alg:exp3p_sparring} as a special case.
\\
\begin{algorithm}[H]
  \algotitle{\texttt{Exp3.P-Sparring}}{alg:exp3p_sparring.title}
  \TitleOfAlgo{\texttt{Exp3.P-Sparring}}
  \KwIn{Time horizon \(T\), Learning rate \(\eta\), Error probability
  \(\delta \in (0, 1)\)} 
  \(\beta = \sqrt{\frac{\log{\frac K \delta}}{TK}},
    \eta = 0.95 \sqrt{\frac{\log K}{TK}},
  \gamma = 1.05 \sqrt{\frac{K \log K}{T}}\) \\
  \(\forall a: \widetilde G_0^A(a) = \widetilde G_0^B(a) = 0\) \\
  \For{\(t = 1, 2, \hdots\)}{
    \(\forall a:
      w_t^A(a) = \exp\left( \eta \widetilde G_{t-1}^A \right),
      w_t^B(a) = \exp\left( \eta \widetilde G_{t-1}^B \right)
    \) \\
    \(\forall a:
      \widetilde p_t^A(a) = (1 - \gamma) \frac{w_t^A(a)}{\sum_{i=1}^K
      w_t^A(i)} + \frac \gamma K,
      \widetilde p_t^B(a) = (1 - \gamma) \frac{w_t^B(a)}{\sum_{i=1}^K
      w_t^B(i)} + \frac \gamma K
    \) \\
    \(\forall a: p_t^A(a) = \frac{\widetilde p_t^A(a)}{\sum_{i=1}^K
      \widetilde p_t^A(i)}, p_t^B(a) = \frac{\widetilde
      p_t^B(a)}{\sum_{i=1}^K \widetilde p_t^B(i)}
    \) \\
    Sample \(A_t \sim p_t^A, B_t \sim p_t^B\) \\
    Play \( (A_t, B_t) \) and observe \(Y_t\) \\
    \(g_t = \frac{Y_t + 1} 2\) \\
    \(\forall a:
      \widetilde G_t^A = \widetilde G_t^A + \frac{g_t \mathbbm 1(a = A_t) +
      \beta}{p_t^A(a)},
      \widetilde G_t^B = \widetilde G_t^B + \frac{(1 - g_t) \mathbbm 1(a =
      B_t) + \beta}{p_t^B(a)}
    \)
  }
  \caption[\texttt{Exp3.P-Sparring}]{\texttt{Exp3.P-Sparring}, based on
    \texttt{Sparring-Exp4.P}
    \citep{pmlr-v40-Dudik15} using \texttt{Exp3.P} implementation and
    parameterisation by \cite{MAL-024}}
  \label{alg:exp3p_sparring}
\end{algorithm}

\section{\nameref{alg:vn_unifkminus1}}
\label{sec:alg_vn_unifkminus1}
We include Algorithm~\ref{alg:vn_unifkminus1} as a heuristic used in
experiments around the von-Neumann winner in the stochastic setting. The
first action \(A_t\) is sampled from the von-Neumann winner of the estimate
\(\widetilde{\mathbf P}^t\) of the cumulative outcome matrix \(\sum_{s=1}^t
\mathbf M^t\), which can be obtained by solving the associated convex
optimisation problem. Similarly to \nameref{alg:exp3_unifkminus1}, the
second action \(B_t\) is sampled uniformly from the remaining actions. After
observing the outcome of the duel between \(A_t\) and \(B_t\),
\(\widetilde{\mathbf P}^{t+1}\) is computed.
\\
\begin{algorithm}[H]
  \algotitle{\texttt{VN+UnifK-1}}{alg:vn_unifkminus1.title}
  \TitleOfAlgo{\texttt{VN+UnifK-1}}
  \(\widetilde{\mathbf P}^0 = \mathbf 0^{K \times K}\) \\
  \For{\(t = 1, 2, \dots\)}{
    Determine \(\mathbf u_t = \arg \max_{\mathbf u \in \Delta^K} \min_j
    \left[ \mathbf u^T \widetilde{\mathbf P}^{t-1} \right]_j \) \\
    Sample \(A_t \sim \mathbf u_t\) \\
    Sample \(B_t\) uniformly from remaining elements \([K] \setminus \left\{
    A_t \right\}\) \\
    Play \( (A_t, B_t) \) and observe \(Y_t\) \\
    \( \forall  i,j \in [K] \times [K]:
      \widetilde P_{ij}^t = \begin{cases}
        \widetilde P_{ij}^{t-1} + \frac{Y_t}{ (K - 1) p_t^{A_t} } \text{ if
        } A_t = i \land B_t = j \\
        \widetilde P_{ij}^{t-1} - \frac{Y_t}{ (K - 1) p_t^{A_t} } \text{ if
        } A_t = j \land B_t = i \\
        \widetilde P_{ij}^{t-1} \text{ otherwise} 
    \end{cases}\)
  }
  \caption[\texttt{VN+UnifK-1}]{\texttt{VN+UnifK-1}, a deterministic
    approximation of the von-Neumann winner of the approximated cumulative
  outcome matrix}
  \label{alg:vn_unifkminus1}
\end{algorithm}

\chapter{Theoretical Results}
\label{ch:theoretical_results}
This chapter describes our theoretical results for
\nameref{alg:exp3_unifkminus1} in both the utility-based setting and the
Borda setting. Furthermore we examine the relation between the utility-based
setting and the non-utility-based setting using different definitions of the
winner.
\section{Non-Stochastic Utility-Based Setting}
\label{sec:theory_adv_util}
This section covers the theoretical analysis of the expected weak regret of
Algorithm~\ref{alg:exp3_unifkminus1} in the non-stochastic
utility-based setting.
\begin{theorem_exp3_unifkminus1_util}
  Given a finite time-horizon \(T\), for \(\eta = \frac 4 K \sqrt{\frac {(K
  - 1) \log K}{3T}}\), Algorithm~\ref{alg:exp3_unifkminus1} satisfies:
  \begin{equation*}
    \mathbb E \left[ R_W^U(T) \right] \leq \sqrt{3 (K-1) T \log K}.
  \end{equation*}
\end{theorem_exp3_unifkminus1_util}
\begin{proof}
Let \(\hist{t}\) denote the set of random variables \( \left\{
\histfull{t} \right\}\). Every round \(t = 1,
\hdots T\), an \texttt{Exp3}-based algorithm picks an action \(A_t\), while
the second action is sampled uniformly at random from the remaining \( K - 1
\) actions. After sampling and observing the outcome \(Y_t\), the algorithm
receives loss
\begin{equation*}
  \frac{1 - Y_t} 2 \in \left\{ 0,1 \right\}.
\end{equation*}
This yields the following loss estimator:
\begin{equation*}
  \lta = \frac{\mathbbm 1 \left( a = A_t \right)}{2 p_t^{A_t}} (1 - Y_t) \in
  \left\{ 0, \frac 1 {p_t^{A_t}} \right\},
\end{equation*}
leading to the following cumulative loss estimator:
\begin{equation*}
  \widetilde L_t(a) = \sum_{s = 1}^t \widetilde{\ell}_s^a
\end{equation*}
with the following first and second moments:
\begin{align*}
  \mathbb E\left[ \lta \middle| \hist{t-1} \right]
  &= \mathbb E\left[ \sum_{i=1}^K p_t^i \sum_{j=1, i \neq j}^K \frac 1 {K-1}
    \frac{\mathbbm 1 \left( a = i \right)}{2 p_t^i} \left(1 - \mathbb E
  \left[ Y_t \middle| A_t = i, B_t = j \right] \right) \middle| \hist{t-1}
\right] \\
  &= \frac 1 {2 (K-1)} \sum_{j=1,j \neq a}^K \left( 1 - \mathbb E \left[ Y_t
  \middle| A_t = a, B_t = j \right] \right) \\
  &= \frac 1 2 + \frac 1 {2 (K-1)} \sum_{j=1}^K \left( \ell_t^a - \ell_t^j
  \right)
  \\ &= \frac 1 2 + \frac K {2 (K-1)} \ell_t^a - \frac 1 {2(K-1)} \sum_{j=1}^K
  \ell_t^j.
\end{align*}
\begin{align*}
  \mathbb E\left[ (\lta)^2 \middle| \hist{t-1} \right]
  &= \mathbb E\left[ \sum_{i=1}^K p_t^i \sum_{j=1, i \neq j}^K \frac 1 {K-1}
    \frac{\mathbbm 1 \left( a = i \right)}{4(p_t^i)^2} (\ell_t^i - \ell_t^j
    + 1)^2 
  \middle| \hist{t-1} \right] \\
  &= \frac 1 {4(K-1)} \mathbb E\left[ \frac 1 {p_t^a} \middle| \hist{t-1}
  \right] \left( \sum_{j=1}^K \left( \ell_t^a - \ell_t^j + 1 \right)^2 - 1
  \right).
\end{align*}
As we are using the loss-based variant of \texttt{Exp3}, the first action is
sampled from the following distribution:
\begin{equation*}
  p_t^a = \frac{\exp\left( -\eta \widetilde L_{t-1}(a) \right)}
  {\sum_{j=1}^K \exp \left( -\eta \widetilde L_{t-1}(j) \right)}.
\end{equation*}
As the learning rate \(\eta\) is positive and the instantaneous loss
estimator is non-negative, the following holds for all actions \(a \in
[K]\)
\citep[Lemma~7]{Seldin:2014:OPA:3044805.3045036}:
\begin{equation}
  \sum_{t=1}^T \sum_{i=1}^K \lti p_t^i - \widetilde L_T(a) \leq \frac{\log
  K}{\eta} + \frac{\eta}{2} \sum_{t=1}^T \sum_{i=1}^K \left( \lti \right)^2
  p_t^i.
  \label{eq:W_combined}
\end{equation}
Taking the expectation with regards to the randomisation of the algorithm
and the sampling of the outcomes:
\begin{equation*}
  \mathbb E\left[ \sum_{t=1}^T \sum_{i=1}^K p_t^i \lti \right]
  = \frac T 2 + \frac K {2(K-1)} \mathbb E\left[ \sum_{t=1}^T \ell_t^{A_t}
  \right] - \frac 1 {2(K-1)} \sum_{t=1}^T \sum_{j=1}^K \ell_t^j.
\end{equation*}
\begin{equation*}
  \mathbb E\left[ \sum_{t=1}^T \lta \right]
  = \frac T 2 + \frac K {2(K-1)} \sum_{t=1}^T \ell_t^a - \frac 1 {2(K-1)}
  \sum_{t=1}^T \sum_{j=1}^K \ell_t^j.
\end{equation*}
For the last part we will use the following technical lemma, whose proof is
provided in Appendix~\ref{app:proof_lemma_technical}.
\begin{lemma_technical}
  \label{lemma:technical}
  Let \(\mathbf x \in [0,1]^n\) with \(n \in \mathbb N\). Then
  \begin{equation*}
    \sum_{i=1}^n \sum_{j=1}^n (x_i - x_j + 1)^2 \leq \frac{3
    n^2} 2.
  \end{equation*}
\end{lemma_technical}
Bounding the expectation of the last term of \eqref{eq:W_combined}:
\begin{align*}
  \mathbb E\left[ \frac \eta 2 \sum_{t=1}^T \sum_{i=1}^K p_t^i \left( \lti
  \right)^2 \right]
  &= \frac \eta {8 (K-1)} \sum_{t=1}^T \mathbb E \left[ \sum_{t=1}^K
  \sum_{i=1}^K p_t^i \mathbb E\left[ \frac 1 {p_t^i} \middle| \hist{t-1}
\right] \left( \sum_{j=1}^K \left( \ell_t^i - \ell_t^j + 1 \right)^2 - 1
\right) \right] \\
&= \frac \eta {8 (K-1)} \sum_{t=1}^T \sum_{i=1}^K \left( \sum_{j=1}^K \left(
\ell_t^i - \ell_t^j + 1 \right)^2 - 1 \right) \\
&\leq \frac{\eta T}{8 (K-1)} \left(\frac{3 K^2} 2 - K \right)
\tag{by Lemma~\ref{lemma:technical}} \\
&= \frac{\eta KT}{8 (K-1)} \left( \frac 3 2 K - 1\right).
\end{align*}
Putting everything together:
\begin{equation*}
  \frac K {2(K-1)} \left( \mathbb E \left[ \sum_{t=1}^T \ell_t^{A_t} \right]
  - \sum_{t=1}^T \ell_t^a \right) \leq \frac{\log K} \eta + \frac{\eta KT}{8
  (K-1)} \left(\frac 3 2 K - 1 \right),
\end{equation*}
which is equivalent to
\begin{equation}
  \mathbb E \left[ \sum_{t=1}^T \ell_t^{A_t} \right] - \sum_{t=1}^T \ell_t^a
  \leq \frac{2 (K - 1) \log K}{K \eta} + \frac{\eta T} 4 \left(\frac 3 2 K -
  1 \right).
  \label{eq:bound2_util}
\end{equation}
The right-hand side of \eqref{eq:bound2_util} is minimised by
\begin{equation*}
  \eta^* = 4 \sqrt{\frac{(K-1) \log K}{(3K - 2) K T}},
\end{equation*}
leading to the following bound:
\begin{align*}
  \mathbb E \left[ \sum_{t=1}^T \ell_t^{A_t} \right] - \sum_{t=1}^T \ell_t^a
  &\leq \frac 1 K \sqrt{(3K - 2) (K - 1) K T \log K} \\
  &\leq \sqrt{3 (K - 1) T \log K}.
\end{align*}
Alternatively, \eqref{eq:bound2_util} can be loosened to
\begin{equation*}
  \mathbb E \left[ \sum_{t=1}^T \ell_t^{A_t} \right] - \sum_{t=1}^T \ell_t^a
  \leq \frac{2 (K - 1) \log K}{K \eta} + \frac{3\eta K T} 8,
\end{equation*}
whose right-hand side is minimised by
\begin{equation*}
  \eta^* = \frac 4 K \sqrt{\frac{(K-1) \log K}{3T}}.
\end{equation*}
This leads to the same regret bound:
\begin{align}
  \mathbb E \left[ \sum_{t=1}^T \ell_t^{A_t} \right] - \sum_{t=1}^T \ell_t^a
  &\leq \sqrt{3 (K - 1) T \log K}.
  \label{eq:bound5_util}
\end{align}
\end{proof}

\section{Non-Stochastic Non-Utility-Based Setting}
\label{sec:theory_adv_nonutil}
This section covers the theoretical analysis of the expected weak regret of
Algorithm~\ref{alg:exp3_unifkminus1} in the non-utility-based setting, with
losses induced by the Borda winner.
\begin{theorem_exp3_unifkminus1_nonutil}
  Given a finite time-horizon \(T\), for \(\eta = 2 \sqrt{\frac{\log
  K}{KT}}\), Algorithm~\ref{alg:exp3_unifkminus1} satisfies:
  \begin{equation*}
    \mathbb E \left[ R_W^N(T) \right] \leq \frac {K-1} K \sqrt{T K \log K}.
  \end{equation*}
\end{theorem_exp3_unifkminus1_nonutil}
\begin{proof}
  Every round \(t = 1, \hdots T\), an \texttt{Exp3}-based algorithm picks an action
\(A_t\), while the second action is sampled uniformly at random from the
remaining \( K - 1 \) actions. After observing the outcome \( M_{A_t B_t}^t
\), the \texttt{Exp3}-based algorithm receives loss
\begin{equation*}
  \frac{1 - M_{A_t B_t}^t} 2 = \frac{M_{B_t A_t}^t + 1} 2 \in \left\{ 0,1
  \right\}.
\end{equation*}
This yields the following loss estimator:
\begin{equation*}
  \lta = \frac{\mathbbm 1 \left( a = A_t \right)}{2 p_t^{A_t}} (M_{B_t
  A_t}^t + 1) \in \left\{ 0, \frac 1 {p_t^{A_t}} \right\},
\end{equation*}
leading to the following cumulative loss estimator:
\begin{equation*}
  \widetilde L_t(a) = \sum_{s = 1}^t \widetilde{\ell}_s^a
\end{equation*}
with the following first and second moments:
\begin{align*}
  \mathbb E\left[ \lta \middle| \hist{t-1} \right]
  &= \mathbb E\left[ \sum_{i=1}^K p_t^i \sum_{j=1, i \neq j}^K \frac 1 {K-1}
    \frac{\mathbbm 1 \left( a = i \right)}{2 p_t^i} (M_{ji}^t + 1) \middle|
    \hist{t-1} \right] \\
  &= \frac 1 {K-1} \sum_{j=1, j \neq a}^K \frac{M_{ja}^t + 1} 2 \\
  &= \frac 1 {K-1} \left( \sum_{j=1}^K \frac{M_{ja}^t + 1} 2 - \frac 1 2
  \right) \\
  &= \frac 1 {K-1} \left( K \ell_t^a - \frac 1 2 \right)
  \tag{by~\ref{eq:def_loss_borda}} \\
  &= \frac K {K-1} \ell_t^a - \frac 1 {2 (K-1)}.
\end{align*}
\begin{align*}
  \mathbb E\left[ (\lta)^2 \middle| \hist{t-1} \right]
  &= \mathbb E\left[ \sum_{i=1}^K p_t^i \sum_{j=1, i \neq j}^K \frac 1 {K-1}
    \frac{\mathbbm 1 \left( a = i \right)}{4 (p_t^i)^2} (M_{ji}^t + 1)^2 
  \middle| \hist{t-1} \right] \\
  &= \frac 1 {4 (K-1)} \mathbb E\left[ \frac 1 {p_t^a} \middle| \hist{t-1}
  \right] \sum_{j=1,j \neq a}^K (M_{ja}^t + 1)^2.
\end{align*}
As we are using the loss-based variant of \texttt{Exp3}, the first action is sampled
from the following distribution:
\begin{equation*}
  p_t^a = \frac{\exp\left( -\eta \widetilde L_{t-1}(a) \right)}
  {\sum_{j=1}^K \exp \left( -\eta \widetilde L_{t-1}(j) \right)}.
\end{equation*}
As before, \eqref{eq:W_combined} holds \(\forall a \in [K]\), due to
\(\eta\) being positive and the instantaneous loss estimators being
non-negative.
Taking the expectation with regards to the randomisation of the algorithm
component-wise:
\begin{align*}
  \mathbb E\left[ \sum_{t=1}^T \sum_{i=1}^K \lti p_t^i \right]
  &= \mathbb E\left[ \sum_{t=1}^T \mathbb E_{i \sim p_t} \left[ \frac K
      {K-1} \ell_t^i - \frac 1 {2 (K-1)} \right] \right] \\
      &= \frac K {K-1} \mathbb E \left[ \sum_{t=1}^T \mathbb E_{i \sim p_t}
      \left[ \ell_t^i \right]\right] - \frac T {2 (K-1)}.
\end{align*}
The second term:
\begin{align*}
  \mathbb E\left[ \sum_{t=1}^T \lta \right]
  &= \mathbb E\left[ \sum_{t=1}^T \mathbb E\left[ \lta \right] \right] \\
  &= \mathbb E\left[ \sum_{t=1}^T \left( \frac K {K-1} \ell_t^a - \frac 1
  {2(K-1)} \right) \right] \\
  &= \frac K {K-1} \sum_{t=1}^T \ell_t^a - \frac T {2 (K-1)}.
\end{align*}
The third term:
\begin{align*}
  \mathbb E\left[ \frac \eta 2 \sum_{t=1}^T \sum_{i=1}^K \left(\lti\right)^2
  p_t^i \right]
  &= \frac \eta 2 \mathbb E\left[ \sum_{t=1}^T \sum_{i=1}^K \mathbb E \left[
  \left(\lti\right)^2 \middle| \hist{t-1} \right] p_t^i\right] \\
  &= \frac \eta 2 \mathbb E\left[ \sum_{t=1}^K \sum_{i=1}^K \left(\frac 1 {4
  (K-1)} \sum_{j=1,j \neq i}^K \left(M_{ja}^t + 1\right)^2 \mathbb E\left[
  \frac 1 {p_t^i} \middle| \hist{t-1}  \right] p_t^i \right) \right] \\
  &= \frac \eta {8 (K-1)} \sum_{t=1}^T \sum_{i=1}^K \sum_{j=1, j \neq i}^K
  \left(M_{ja}^t + 1\right)^2 \\
  &\leq \frac \eta {8 (K-1)} \sum_{t=1}^T \frac{4 K (K-1)} 2 \\
  &= \frac 1 4 \eta T K.
\end{align*}
Putting everything together:
\begin{equation*}
  \frac K {K-1} \left( \mathbb E\left[ \sum_{t=1}^T \ell_t^{A_t} \right] -
  \sum_{t=1}^T \ell_t^a \right) \leq \frac{\log K} \eta + \frac 1 4 \eta T K,
\end{equation*}
which is equivalent to
\begin{equation*}
  \mathbb E\left[ \sum_{t=1}^T \ell_t^{A_t} \right] - \sum_{t=1}^T \ell_t^a
  \leq \frac{(K - 1) \log K} {K \eta} + \frac 1 4 \eta T (K - 1).
\end{equation*}
Assuming that \(\eta > 0\), the right-hand side of \eqref{eq:bound2_util} is
minimised by
\begin{equation*}
  \eta^* = 2 \sqrt{\frac{\log K}{KT}},
\end{equation*}
which leads to
\begin{equation}
  \sum_{t=1}^T \mathbb E \left[ \ell_t^{A_t} \right] - \sum_{t=1}^T \ell_t^a
  \leq \frac{K-1} K \sqrt{T K \log K}.
  \label{eq:bound3_nonutil}
\end{equation}
\end{proof}

\section{Relation between Non-Utility-Based Regret and Utility-Based Regret}
\label{sec:theory_relation_nonutil_util}
To distinguish between the utility-based loss and the non-utility-based
loss, let \(\bar \ell_t^i\) denote the utility-based loss, as defined in
\eqref{eq:def_utility_loss}. We will show that in expectation with respect
to the sampling process of the outcomes, the Borda winner, the Copeland
winner, and the von-Neumann winner are identical to the utility-based
winner, and relate their induced regret to the equivalent utility-based
regret.

\subsection{Borda Winner}
\label{subsec:theory_relation_borda}
Assuming a utility-based setting and a linear function, we can use
\eqref{eq:bt_pr}, reflecting our updated notation: 
\begin{equation}
  \mathbb E\left[ M_{ij}^t \right] = \bar \ell_t^i - \bar \ell_t^j.
  \label{eq:alt_EMij}
\end{equation}
Examining the Borda loss \eqref{eq:def_loss_borda} in expectation yields
\begin{align*}
  \mathbb E \left[ \ell_t^i \right]
  &= \mathbb E \left[ \frac 1 2 + \frac 1 {2K} \sum_{j=1}^K M_{ij}^t \right]
  \\ &= \frac 1 2 + \frac 1 {2K} \sum_{j=1}^K \left( \bar \ell_t^i - \bar
  \ell_t^j \right) \\
  &= \frac 1 2 + \frac 1 2 \bar \ell_t^i - \frac 1 {2K} \sum_{j=1}^K \bar
  \ell_t^j.
\end{align*}
Given a pair of sequences of actions \( \left( A_t \right)_{t=1}^T, \left(
B_t \right)_{t=1}^T\), this translates into the following non-utility-based
expected weak regret, with respect to the randomness of the sampling
process:
\begin{align*}
  \mathbb E \left[ R_W^N(T) \right] 
  &= \sum_{t=1}^T \mathbb E \left[ \min\left\{ \ell_t^{A_t}, \ell_t^{B_t}
    \right\} \right]- \sum_{t=1}^T \mathbb E \left[ \ell_t^{a^*} \right] \\
  &\leq \sum_{t=1}^T \min\left\{ \mathbb E \left[ \ell_t^{A_t} \right], 
  \mathbb E \left[ \ell_t^{B_t} \right] \right\} - \sum_{t=1}^T \mathbb E
  \left[ \ell_t^{a^*} \right] \\
  &= \frac T 2 + \frac 1 2 \sum_{t=1}^T \min \left\{ \bar \ell_t^{A_t}, \bar
  \ell_t^{B_t} \right\} - \frac 1 {2K} \sum_{t=1}^T \sum_{j=1}^K \bar \ell_t^j
  - \frac T 2 - \frac 1 2 \sum_{t=1}^T \bar \ell_t^{a^*} + \frac 1 {2K}
  \sum_{t=1}^T \sum_{j=1}^K \bar \ell_t^j \\
  &= \frac 1 2 \sum_{t=1}^T \min \left\{ \bar \ell_t^{A_t}, \bar \ell_t^{B_t}
  \right\} - \frac 1 2 \sum_{t=1}^T \bar \ell_t^{a^*} \\
  &= \frac 1 2 R_W^U(T).
\end{align*}
This is identical to the result in \cite{Gajane:2015:REW:3045118.3045143},
who showed that utility-based regret is twice the Condorcet regret assumed
by \cite{Yue:2012:KDB:2240304.2240501}.

Using \eqref{eq:bound5_util}, and assuming \(K > 4\), we obtain a tighter
bound on the expected non-utility-based weak regret, with respect to both
the randomisation of the algorithm and the randomness induced by the
sampling process of the observations:
\begin{align*}
  \mathbb E \left[ R_W^N(T) \right]
  &\leq \frac 1 2 \mathbb E \left[ R_W^U \right] \\
  &= \frac 1 2 \sqrt{3 (K - 1) T \log K},
\end{align*}
improving the result from \eqref{eq:bound3_nonutil} by a factor of \(\frac 2
{\sqrt 3} \sqrt{\frac{K - 1} K}\).

\subsection{Copeland Winner}
\label{subsec:theory_relation_copeland}

We examine the Copeland loss \eqref{eq:def_loss_copeland} in expectation,
using \eqref{eq:alt_EMij}:
\begin{align*}
  \mathbb E \left[ \ell_a \right]
  &= \frac 1 {K - 1} \sum_{j=1}^K \mathbbm 1 \left( \mathbb E \left[
  \sum_{t=1}^T M_{aj}^t \right] < 0 \right) \\
  &= \frac 1 {K - 1} \mathbbm 1 \left( \left( \sum_{t=1}^T  \left( \bar
  \ell_t^j - \bar \ell_t^a \right) \right) < 0  \right) \\
  &= \frac 1 {K - 1} \mathbbm 1 \left( \sum_{t=1}^T \bar \ell_t^j <
  \sum_{t=1}^T \bar \ell_t^a \right) \\
\end{align*}
Due to to the use of the indicator function, the Copeland loss is not
directly relatable to the utility-based loss. We will further more show in
Section~\ref{sec:exp_copeland} that all the algorithms considered in
Chapter~\ref{ch:algorithms} can suffer linear regret even when considering
the stochastic setting.

\subsection{Von-Neumann Winner}
\label{subsec:theory_relation_vonneumann}
Examining the weak von-Neumann regret \eqref{eq:def_regret_von_neumann},
\eqref{eq:def_psi_weak} in expectation using \eqref{eq:alt_EMij}:
\begin{align*}
  \mathbb E \left[ R_W^N(T) \right]
  &= \sum_{t=1}^T \mathbb E \left[\min \left\{ \ell_t^{A_t}, \ell_t^{B_t}
  \right\} \right] \\
    &\leq \sum_{t=1}^T \mathbb \min \left\{ \mathbb E \left[ \ell_t^{A_t}
    \right], \mathbb E \left[ \ell_t^{B_t} \right] \right\} \\
    &= \sum_{t=1}^T \mathbb \min \left\{ \mathbb E \left[ M_{a_V^* A_t}^t
      \right], \mathbb E \left[ M_{a_V^* B_t}^t \right]
    \right\} \\
    &= \sum_{t=1}^T \mathbb \min \left\{ \bar \ell_t^{A_t} - \bar
      \ell_t^{a_V^*}, \bar \ell_t^{B_t} - \bar \ell_t^{a_V^*} \right\} \\
    &= \sum_{t=1}^T \mathbb \min \left\{ \bar \ell_t^{A_t}, \bar \ell_t^{B_t}
    \right\} - \sum_{t=1}^T \bar \ell_t^{a_V^*} \\
    &\leq \sum_{t=1}^T \mathbb \min \left\{ \bar \ell_t^{A_t}, \bar
    \ell_t^{B_t} \right\} - \sum_{t=1}^T \bar \ell_t^{a_U^*} \\
    &= R_W^U(T).
\end{align*}
The last inequality holds because of \eqref{eq:def_best_utility}. This shows
that the expected weak regret induced by the von-Neumann setting is
upper-bounded by the utility-based weak regret. The same holds for the
strong regret, which can be proved analogously.

\chapter{Experimental Results}
\label{ch:experimental_results}
This section evaluates the performance of \nameref{alg:exp3_unifkminus1} in
the stochastic setting experimentally, complementing our theoretical results
in the adversarial setting.

\section{Borda Regret}
\label{sec:exp_borda}
To the best of our knowledge, there exist no other algorithms covering the
adversarial duelling bandits problem in the Borda setting. Due to the
absence of algorithms for comparison and the limited informational value of
standalone experiments examining how an algorithm behaves when the
distribution underlying the outcome generation process changes, we show
only that the other algorithms covered in Chapter~\ref{ch:algorithms} are
not suitable for the Borda setting and focus on the stochastic setting
instead.

\subsection{\nameref{alg:exp3_unifkminus1} in the Stochastic Setting}
\label{subsec:exp_borda_borda}
We modified the simulation framework provided by \cite{pmlr-v40-Komiyama15}
to account for both the Borda loss~\eqref{eq:def_loss_borda} and the weak
regret~\eqref{eq:def_regret_weak}. We extended the collection of algorithms
with an implementation of \texttt{WS-W}, as presented in
\cite[Algorithm~1]{pmlr-v70-chen17c}, as well as a generic Borda reduction
\texttt{UCB+UnifK-1}, which replaces the \texttt{Exp3} algorithm of
\nameref{alg:exp3_unifkminus1} with an \texttt{UCB}-based algorithm.

\subsubsection{Experiment Setup}
As all the other algorithms were designed for the Condorcet winner setting,
we consider only preference matrices which induce a Condorcet winner which
is identical to their unique Borda winner. We excluded any algorithms which
assume the existence of a total order of the actions' preferences.
\begin{description}
  \item[arXiv] This preference matrix over \(K = 6\) actions was derived by
    \cite{Yue:2011:BMB:3104482.3104513} from the data by
    \cite{Radlinski:2008:CDR:1458082.1458092}, who conducted interleaving
    experiments by providing a customised search engine on the arXiv
    preprint repository\footnote{\url{https://arxiv.org/}}. An inconsistency
    in the original preference matrix was solved by decreasing \(P_{24}\)
    and \(P_{42}\) by \(0.01\), yielding
    \begin{equation*}
      \mathbf P = \begin{bmatrix}
        0.5  & 0.55 & 0.55 & 0.54 & 0.61 & 0.61 \\
        0.45 & 0.5  & 0.55 & 0.55 & 0.58 & 0.6 \\
        0.45 & 0.45 & 0.5  & 0.54 & 0.51 & 0.56 \\
        0.46 & 0.45 & 0.46 & 0.5  & 0.46 & 0.5 \\
        0.39 & 0.42 & 0.49 & 0.46 & 0.5  & 0.51 \\
        0.39 & 0.40 & 0.44 & 0.5  & 0.49 & 0.5
      \end{bmatrix}.
    \end{equation*}
  \item[cyclic] This small synthetic preference matrix with \(K = 4\)
    actions has no total order, as the preference between the actions which
    are unequal to the Condorcet winner is not transitive
    \citep{pmlr-v40-Komiyama15}.
    \begin{equation*}
      \mathbf P = \begin{bmatrix}
        0.5 & 0.6 & 0.6 & 0.6 \\
        0.4 & 0.5 & 0.9 & 0.1 \\
        0.4 & 0.1 & 0.5 & 0.9 \\
        0.4 & 0.9 & 0.1 & 0.5
      \end{bmatrix}
    \end{equation*}
  \item[sushi] Based on the SUSHI preference datasets obtained through
    surveys involving assigning scores and ranking different types of sushi
    \citep{Kamishima:2003:NCF:956750.956823}, this preference matrix was
    derived by considering only the \(K = 16\) most popular types of sushi,
    and can be found in
    \cite[Table~3(a)]{Komiyama:2016:CDB:3045390.3045521}.
\end{description}
Using time horizons \(T = 10^4, 10^5\), the weak regret was averaged over
100 iterations. This experiment can be reproduced by following the steps
described in Appendix~\ref{app:exp_stochastic_borda}. The
\texttt{RMED}-based algorithms use the original parameterisation
\citep[Section~4.1]{pmlr-v40-Komiyama15}. \texttt{RUCB} and
\texttt{UCB+UnifK-1} use \(\alpha=0.51\).

\subsubsection{Results}
\begin{figure}[H]
\begin{minipage}{.5\textwidth}
  \includegraphics[width=\textwidth]
  {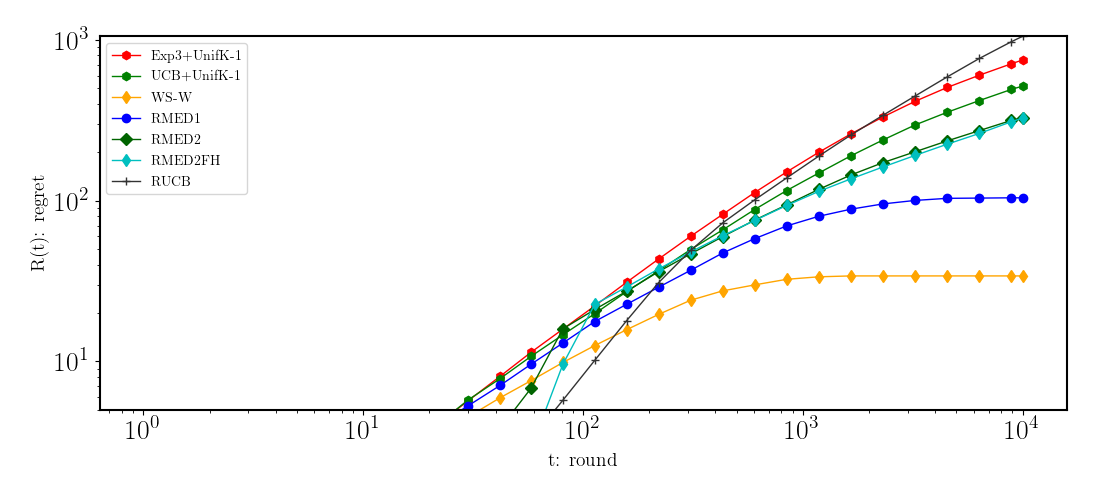}
\end{minipage}%
\begin{minipage}{.5\textwidth}
  \includegraphics[width=\textwidth]
  {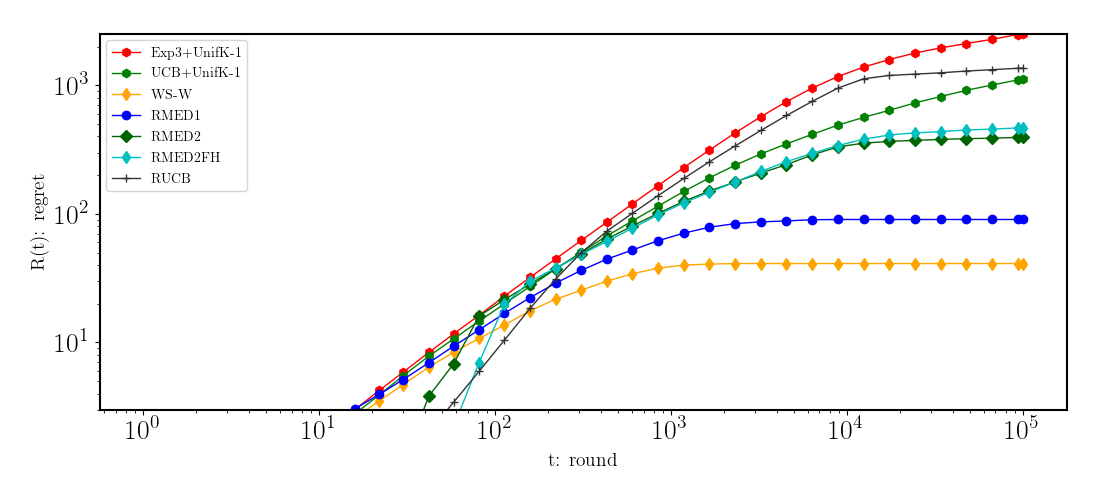}
\end{minipage}
\begin{minipage}{.5\textwidth}
  \includegraphics[width=\textwidth]
  {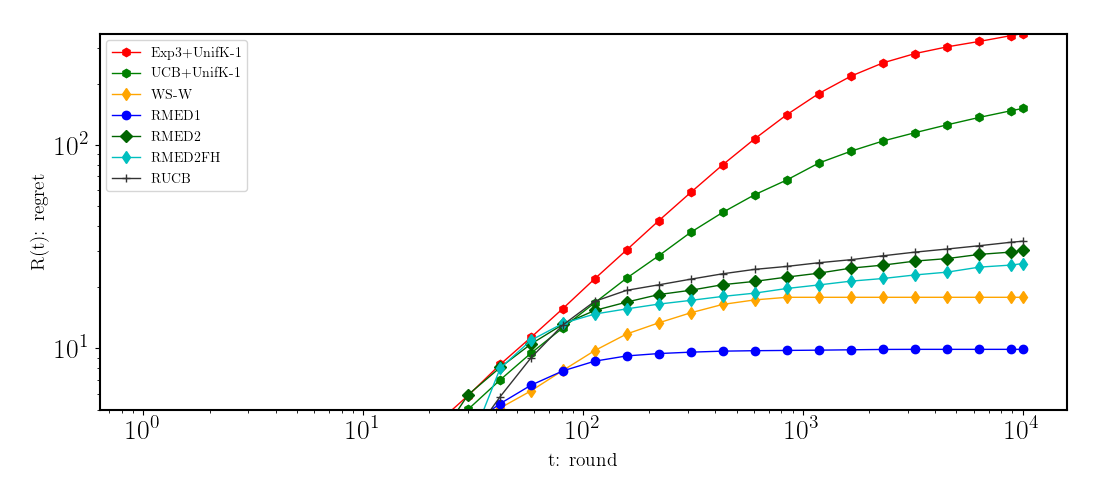}
\end{minipage}%
\begin{minipage}{.5\textwidth}
  \includegraphics[width=\textwidth]
  {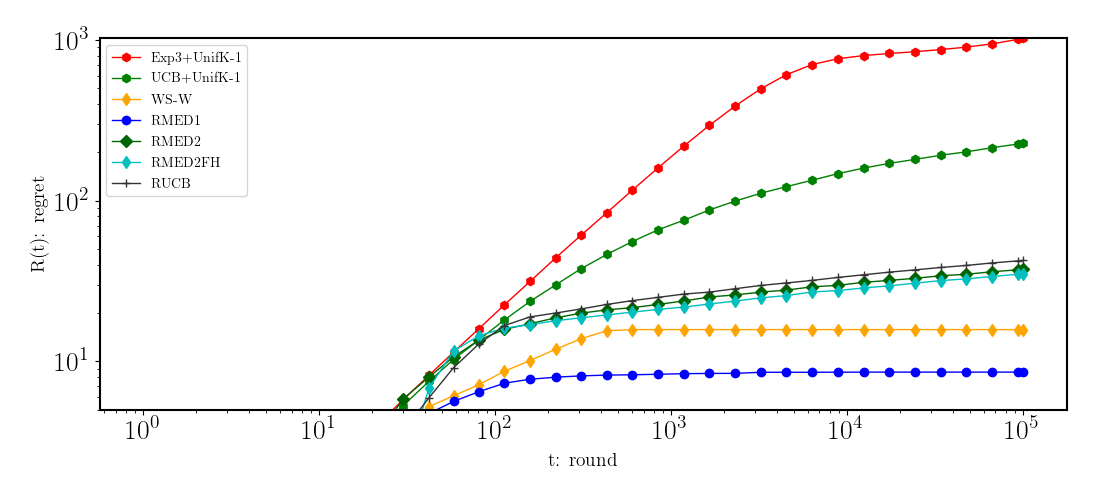}
\end{minipage}
\begin{minipage}{.5\textwidth}
  \includegraphics[width=\textwidth]
  {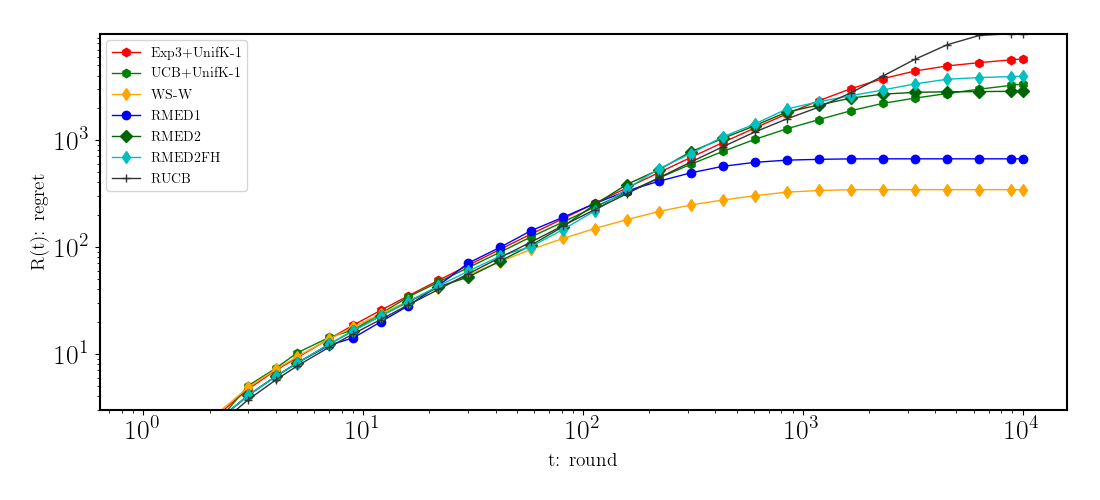}
\end{minipage}%
\begin{minipage}{.5\textwidth}
  \includegraphics[width=\textwidth]
  {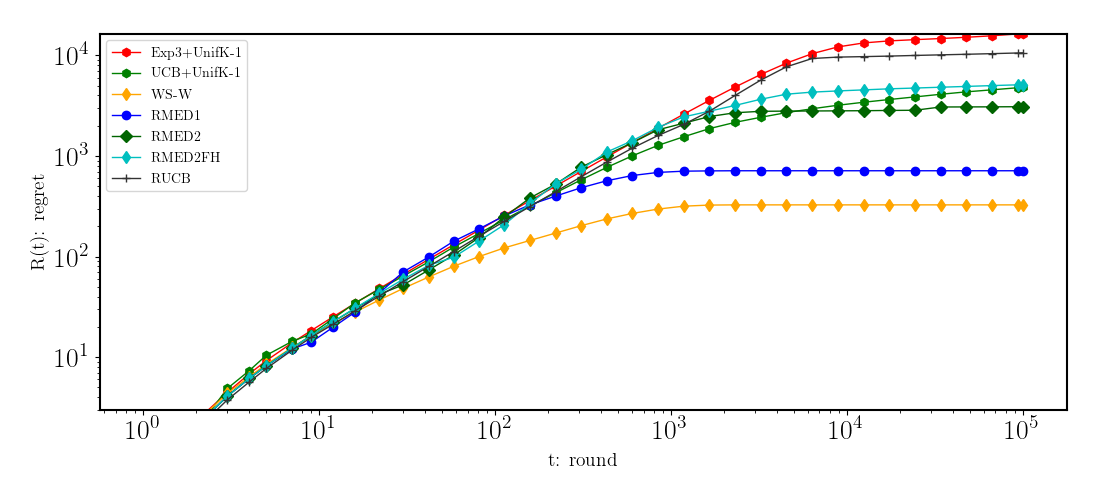}
\end{minipage}
\caption[Weak Regret of \nameref{alg:exp3_unifkminus1} in Stochastic Borda
Setting]{Weak regret of \nameref{alg:exp3_unifkminus1} and other algorithms
in the stochastic Borda setting with \(T=10^4\) (left) and \(T=10^5\)
(right) for \textbf{arXiv} (top), \textbf{cyclic} (middle), and
\textbf{sushi} (bottom)}
\label{fig:exp_borda_stochastic}
\end{figure}

Figure~\ref{fig:exp_borda_stochastic} shows the cumulative weak regret for a
range of algorithms. It becomes obvious that \nameref{alg:exp3_unifkminus1}
is greatly outperformed by the specialised, more recent algorithms
\texttt{RMED} and \texttt{WS-W}, the latter harnessing the knowledge that it
is run in a weak regret setting, yielding regret constant in \(T\).
\texttt{RUCB} performs well on the \textbf{cyclic} dataset, but
performs worse than \nameref{alg:exp3_unifkminus1} on the datasets
\textbf{arXiv} and \textbf{sushi} for the shorter time horizon \(T = 10^4\).
\texttt{UCB+UnifK-1} performs noticeably better than
\nameref{alg:exp3_unifkminus1}, but is unable to keep up with the
best-performing specialised algorithms. 

\subsection{\nameref{alg:exp3_sparring}, \nameref{alg:exp3p_sparring},
\nameref{alg:vn_unifkminus1}}
\label{subsec:exp_borda_vn}
We would like to show the insufficiency of \nameref{alg:exp3_sparring},
\nameref{alg:exp3p_sparring}, and \nameref{alg:vn_unifkminus1} in the Borda
setting. To do so, we generate a sequence of outcomes based on the
preference matrix
\begin{equation*}
  \mathbf P = \begin{bmatrix}
    0.5  & 1.0 & 0.55 & 0.55 & 0.55 \\
    0.0  & 0.5 & 1.0  & 1.0  & 1.0 \\
    0.45 & 0.0 & 0.5  & 0.5  & 0.5 \\
    0.45 & 0.0 & 0.5  & 0.5  & 0.5 \\
    0.45 & 0.0 & 0.5  & 0.5  & 0.5
  \end{bmatrix},
  \label{eq:pref_borda_vn}
\end{equation*}
whose Borda winner is the second action, while its von-Neumann winner has
only support by the first action. Let this sequence be denoted by
\begin{equation}
  \left( \mathbf M^t \right)_{t=1}^T \text{ with } \mathbf M^t \sim \mathbf
  P \text{ such that } \frac 1 T \sum_{t=1}^T \mathbf M^t = \mathbf P.
  \label{eq:seq_borda_vn}
\end{equation}

\subsubsection{Experiment Setup}
Due to \(20\) being the smallest positive number guaranteeing integer
multiplicity, we construct a sequence of \(\tau = 20\) outcome matrices
matrices \( \left( \mathbf M^t \right)_{t=1}^\tau \) the following way: for
every sorted pair of actions \( (i, j) \subseteq [K] \times [K] \) with \( i
< j \), we construct a sequence of binary outcomes by considering a random
permutation of the multiset with support in \(\left\{1, -1\right\}\), with
multiplicities \(\tau \cdot P_{ij}\) and \(\tau \cdot P_{ji}\),
respectively.  Assuming that \(T \equiv 0 \pmod \tau\), repeating the
resulting sequence \( \frac T \tau\) times yields a sequence fulfilling
\eqref{eq:seq_borda_vn} by construction. Sequences were generated for
\(T = 10^3, 10^4, 10^5\), and the execution of all three algorithms was
repeated 100 times, allowing to approximate the mean and standard deviation
with respect to the algorithm's internal randomisation. This experiment can
be reproduced by following the steps described in
Appendix~\ref{app:exp_borda_vn}.

\subsubsection{Results}
\begin{figure}[H]
\begin{minipage}{.5\textwidth}
  \includegraphics[width=\textwidth]{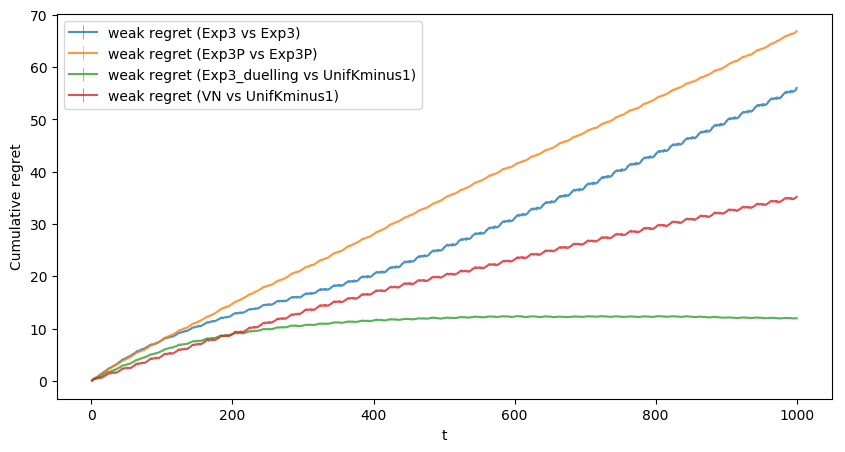}
\end{minipage}%
\begin{minipage}{.5\textwidth}
  \includegraphics[width=\textwidth]{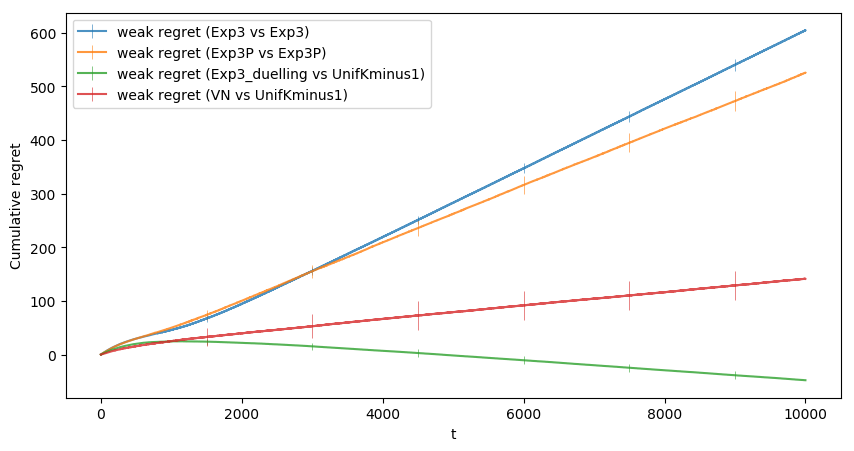}
\end{minipage}
\raggedright
\begin{minipage}{.5\textwidth}
  \includegraphics[width=\textwidth]{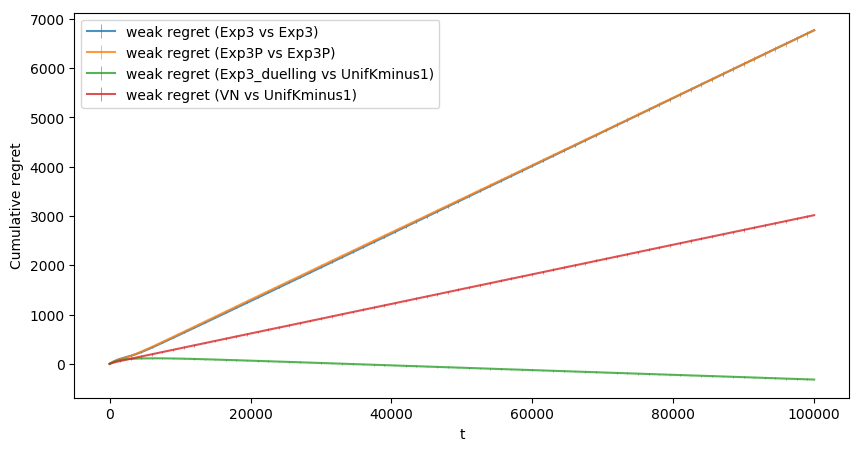}
\end{minipage}
\caption[Weak Regret of \nameref{alg:exp3_unifkminus1},
  \nameref{alg:exp3_sparring}, \nameref{alg:exp3p_sparring}, and
\nameref{alg:vn_unifkminus1} in Borda Setting]
{Weak Regret of \nameref{alg:exp3_unifkminus1},
\nameref{alg:exp3_sparring}, \nameref{alg:exp3p_sparring}, and
\nameref{alg:vn_unifkminus1} with \(T=10^3\) (top-left), \(T=10^4\)
(top-right), and \(T=10^5\) (bottom) in Borda Setting}
    \label{fig:exp_borda_vn}
\end{figure}

Figure~\ref{fig:exp_borda_vn} depicts the mean cumulative weak regret for
all four algorithms. In contrast to \nameref{alg:exp3_unifkminus1}, whose
weak regret behaves sublinearly in \(t\), the weak regret of the other three
algorithms grows linearly in \(t\), suggesting that they are not suitable
for the Borda setting.

\section{Copeland Regret}
\label{sec:exp_copeland}
The analysis of \nameref{alg:exp3_unifkminus1.title} in the
non-utility-based setting in Section~\ref{sec:theory_adv_nonutil} was
restricted to the loss setting induced by the Borda winner. At the same
time, \cite{pmlr-v40-Dudik15} claimed that
Algorithm~\ref{alg:exp3p_sparring} suffers sublinear strong regret in the
von-Neumann setting. This suggests that these algorithms can suffer linear
Copeland regret even in non-adversarial instances, which we show
experimentally.

\subsection{\nameref{alg:exp3_unifkminus1}}
\label{subseq:exp_copeland_borda}
As \nameref{alg:exp3_unifkminus1} spends on expectation at most \(
O\left(\sqrt{K T \log K} \right) \) on exploration, the cumulative
difference between the Copeland loss incurred by the first action, which is
chosen by an instance of the \texttt{Exp3} algorithm, and the equivalent loss
incurred by the Copeland winner can be linear in \(t\) if the Borda winner
is not identical to the Copeland winner. We will assume the following
preference matrix:
\begin{equation*}
  \mathbf P = \begin{bmatrix}
    0.5 & 1.0 & 1.0 & 0.4 & 0.4 \\
    0.0 & 0.5 & 0.6 & 0.6 & 0.6 \\
    0.0 & 0.4 & 0.5 & 0.4 & 0.6 \\
    0.6 & 0.4 & 0.6 & 0.5 & 0.4 \\
    0.6 & 0.4 & 0.4 & 0.6 & 0.5
  \end{bmatrix}.
\end{equation*}
Given a sequence
\begin{equation*}
  \left( \mathbf M^t \right)_{t=1}^T \text{ with } \mathbf M^t \sim \mathbf
  P \text{ such that } \frac 1 T \sum_{t=1}^T \mathbf M^t = \mathbf P,
\end{equation*}
the latter assumption implying that, almost surely, the first action is
constitutes the unique Borda winner, while the Copeland winner has only
support by the second action. The cumulative Borda loss and Copeland loss
vectors, in expectation, are given by
\begin{align*}
  \mathbb E \left[\mathbf L_B \right] &= \frac T K
  \begin{bmatrix}
    1.7 & 2.7 & 3.1 & 2.5 & 2.5
  \end{bmatrix}^T \\
  \mathbb E \left[ \mathbf L_C \right] &= \frac T {K - 1}
  \begin{bmatrix}
    2 & 1 & 3 & 2 & 2
  \end{bmatrix}^T.
\end{align*}
This implies that, on expectation, the Borda winner suffers Copeland loss
\( \frac 2 {K - 1} \) every round, while sampling an action from the
remaining arms uniformly at random suffers Copeland loss \( \frac 2 {K - 1}
\) as well. This implies that \nameref{alg:exp3_unifkminus1} suffers at most
\footnote{As \(\mathbb E \left[ \min \left\{\cdot, \cdot \right\} \right]
\leq \min \left\{ \mathbb E\left[ \cdot \right], \mathbb E\left[ \cdot
\right] \right\}\), preventing the use of the result as a lower bound.}
\( \frac 1 {K - 1} \left( \frac 1 4 \cdot 1 + \frac 3 4 \cdot 2 - 1 \right)
= \frac 3 {16} \) instantaneous weak regret in expectation
whenever it selects the Borda winner as first action. As we are not
interested in upper bounding the expected regret, we will resort to
experimental verification.

\subsubsection{Experiment Setup}
The experiment setup is similar to the one described
in~\ref{subsec:exp_borda_vn}, with \(\tau = 10\), and \(T = 10^3, 10^4\).
The execution of \nameref{alg:exp3_unifkminus1} was repeated 100 times, to
obtain a better estimate of the expected regret with respect to the
algorithm's internal randomisation. This experiment can be reproduced by
following the steps described in
Appendix~\ref{app:exp_copeland_exp3_unifkminus1}.

\subsubsection{Results}
\begin{figure}[H]
  \centering
\begin{minipage}{.5\textwidth}
  \includegraphics[width=\textwidth]
  {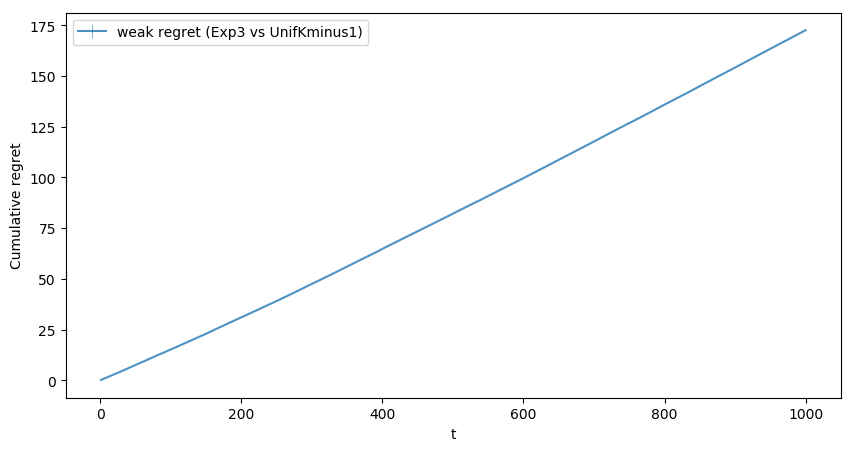}
\end{minipage}%
\begin{minipage}{.5\textwidth}
  \centering
  \includegraphics[width=\textwidth]
  {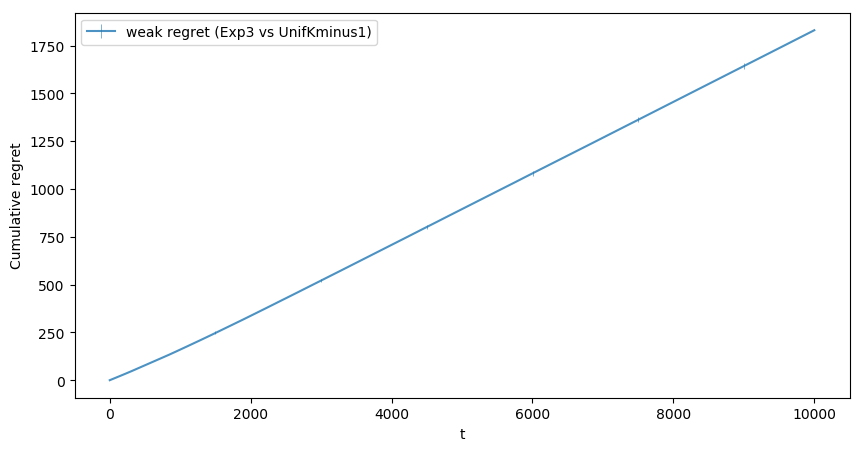}
\end{minipage}
\caption[Weak Regret of \nameref{alg:exp3_unifkminus1} in Copeland
Setting]{Weak Regret of \nameref{alg:exp3_unifkminus1} with \(T=1000\)
(left) and \(T=10000\) (right) in Copeland Setting}
  \label{fig:exp_copeland_exp3_unifkminus1}
\end{figure}

Figure~\ref{fig:exp_copeland_exp3_unifkminus1} depicts the mean weak regret
accumulated over time for the two experiments, which grows linearly
in \(t\) with very little variance. This supports our claim that
\nameref{alg:exp3_unifkminus1} is not suitable in scenarios requiring the
minimisation of the Copeland loss, and can be forced to suffer weak regret
linear in \(t\) if the Copeland winner is not identical to the Borda winner. 

\subsection{\nameref{alg:exp3_sparring}, \nameref{alg:exp3p_sparring},
\nameref{alg:vn_unifkminus1}}
\label{subsec:exp_copeland_vn}
Unlike \nameref{alg:vn_unifkminus1}, which was specifically designed to
approximate the von-Neumann winner, we include \nameref{alg:exp3p_sparring}
in this section due to \cite{pmlr-v40-Dudik15} claiming the very same. We add
\nameref{alg:exp3_sparring} for completeness. Assuming that all these focus
on finding the von-Neumann winner, we conjecture that they can suffer linear
weak regret in the Copeland setting, given that the von-Neumann winner has
support by actions which do not constitute the Copeland winner. 
\cite[p.17]{pmlr-v40-Dudik15} used the following preference matrix for
illustrating a stochastic non-utility-based case inducing a von-Neumann with
no support of the Copeland winner \footnote{The elements of the preference
matrix suggested by \cite{pmlr-v40-Dudik15} do not denote the bias of the
Bernoulli random variable inducing the individual outcomes. Given
their preference matrix \(\mathbf P'\), the result
in~\eqref{eq:pref_copeland_vn} was obtained by computing \(\mathbf P =
\mathbf P' / 2 + 0.5\).}:
\begin{equation}
  \mathbf P = \begin{bmatrix}
    0.5   & 0.75 & 0.25 & 0.75 & 0.025 \\
    0.25  & 0.5  & 0.75 & 0.4  & 0.75 \\
    0.75  & 0.25 & 0.5  & 0.4  & 0.75 \\
    0.25  & 0.6  & 0.6  & 0.5  & 0.75 \\
    0.975 & 0.25 & 0.25 & 0.25 & 0.5
  \end{bmatrix},
  \label{eq:pref_copeland_vn}
\end{equation}
which induces a von-Neumann winner which is a uniform distribution over the
first
three actions \(\begin{bmatrix} 1/3 & 1/3 & 1/3 & 0 & 0 \end{bmatrix}\), and
a Copeland winner focussing on the fourth action \citep{pmlr-v40-Dudik15}. 

\subsubsection{Experiment Setup}
The experiment setup is similar to the one described in
\ref{subsec:exp_copeland_vn} with \(\tau = 40\), and \(T = 10^3, 10^4\).
Again, all algorithms were run 100 times. This experiment can be reproduced
by following the steps described in Appendix~\ref{app:exp_copeland_vn}.

\subsubsection{Results}
\begin{figure}[H]
  \centering
\begin{minipage}{.5\textwidth}
  \centering
  \includegraphics[width=\textwidth]{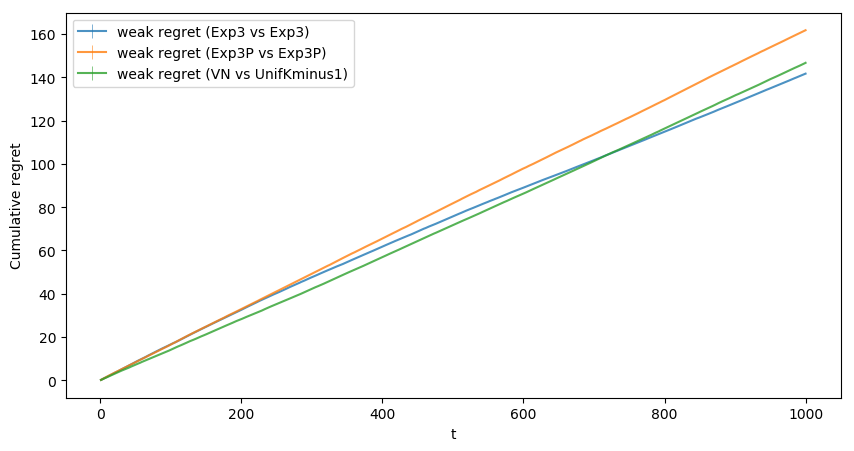}
\end{minipage}%
\begin{minipage}{.5\textwidth}
  \centering
  \includegraphics[width=\textwidth]{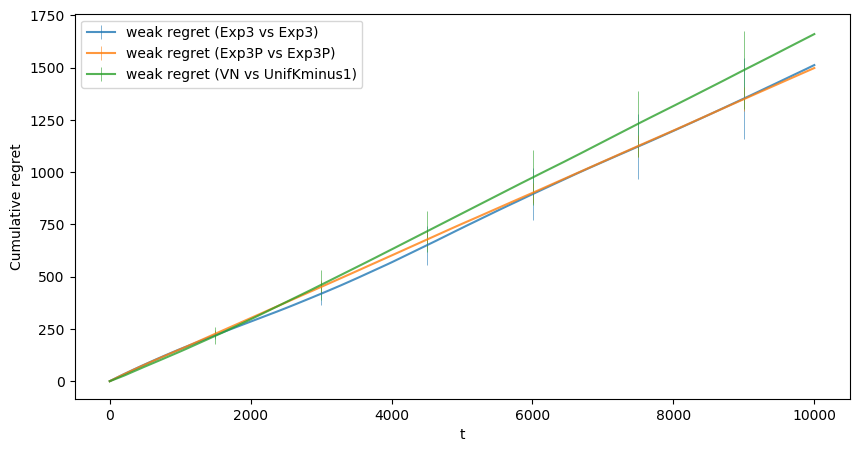}
\end{minipage}
\caption[Weak Regret of \nameref{alg:exp3_sparring},
\nameref{alg:exp3p_sparring}, and \nameref{alg:vn_unifkminus1} in Copeland
Setting]{Weak Regret of \nameref{alg:exp3_sparring},
  \nameref{alg:exp3p_sparring}, and \nameref{alg:vn_unifkminus1} with
\(T=10^3\) (left) and \(T=10^4\) (right) in Copeland Setting}
  \label{fig:exp_copeland_vn}
\end{figure}

Figure~\ref{fig:exp_copeland_vn} depicts the mean weak regret of the three
algorithms taken into consideration in this experiment. The mean weak regret
increases linearly in \(t\), confirming our previous claim. In conclusion,
none of the algorithms suggested in Chapter~\ref{ch:algorithms} are
sufficient for the non-utility-based duelling bandit problem in the Copeland
setting.

\section{Von-Neumann Regret}
\label{sec:exp_vn}

This section shows that \nameref{alg:exp3_unifkminus1} is not suitable for
the von-Neumann setting. The preference matrix \eqref{eq:pref_borda_vn}
induces a von-Neumann winner which has no support by the Borda winner.
However, the uniform exploration behaviour guarantees that in any round,
\nameref{alg:exp3_unifkminus1} suffers on expectation at most
\begin{equation*}
  1 - \left( \frac 3 4 \cdot 0.45 + \frac 1 4 0.5 \right) = 0.5375
\end{equation*}
instantaneous loss, which is close to the von-Neumann winner's loss of
\(0.5\) when considering weak regret. This makes it more difficult to
observe high regret in experiments
with reasonably small time horizons. For this reason we introduce a larger
preference matrix with \(K = 16\) arms. Based on the preference matrix
\eqref{eq:pref_copeland_vn}, we created a larger matrix with \(K = 16\)
arms, which increases the loss of a uniformly sampled action without
modifying the Borda winner or the von-Neumann winner.
\begin{equation*}
  \mathbf P = \begin{bmatrix}
    0.5   & 0.75 & 0.25 & 1.0  & 0.025 & 0.8 & \hdots & 0.8 \\
    0.25  & 0.5  & 0.75 & 1.0  & 0.75  & 0.8 & \hdots & 0.8 \\
    0.75  & 0.25 & 0.5  & 1.0  & 0.75  & 0.8 & \hdots & 0.8 \\
    0.0   & 0.0  & 0.0  & 0.5  & 0.75  & 1.0 & \hdots & 1.0 \\
    0.975 & 0.25 & 0.25 & 0.25 & 0.5   & 0.8 & \hdots & 0.8 \\
    0.2   & 0.2  & 0.2  & 0.0  & 0.2   & 0.5 & \hdots & 0.5 \\
    \vdots&\vdots&\vdots&\vdots&\vdots&\vdots& \ddots & \vdots \\
    0.2   & 0.2  & 0.2  & 0.0  & 0.2   & 0.5 & \hdots & 0.5 \\
  \end{bmatrix}
\end{equation*}
Similar to the original setting, the von-Neumann winner is \(\begin{bmatrix}
1/3 & 1/3 & 1/3 & 0 & \hdots & 0 \end{bmatrix}\), while the unique Borda
winner is the fourth action.

\subsection{Experiment Setup}
This experiment setup is similar to the one described in
Subsection~\ref{subsec:exp_copeland_vn}, using \(\tau = 40\), \(T = 10^3,
10^4, 10^5\), and 10 iterations. This experiment can be reproduced by
following the steps described in Appendix~\ref{app:exp_vn}.

\subsection{Results}
Figure~\ref{fig:exp_vn} depicts the mean weak regret of the four
algorithms taken into consideration in this experiment. The regret of the
\nameref{alg:exp3_sparring} and \nameref{alg:vn_unifkminus1} quickly becomes
negative. \nameref{alg:exp3p_sparring} has noticeably more exploration,
suggesting that the \( O\left(\sqrt{KT \log{\frac K \delta}} \right) \)
bound on the expected strong regret, as argued for by
\cite{pmlr-v40-Dudik15}, hides a large constant. The algorithm of interest,
\nameref{alg:exp3_unifkminus1}, suffers linear regret in this experiment, as
its first arm approximates the wrong winner without its second arm
compensating for it. This shows that it is not suitable for the von-Neumann
setting.

\begin{figure}[H]
  \centering
\begin{minipage}{.5\textwidth}
  \centering
  \includegraphics[width=\textwidth]{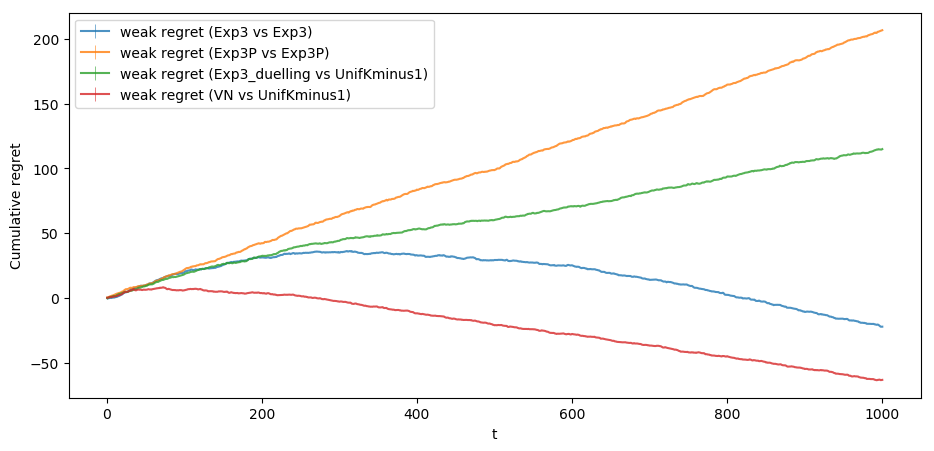}
\end{minipage}%
\begin{minipage}{.5\textwidth}
  \centering
  \includegraphics[width=\textwidth]{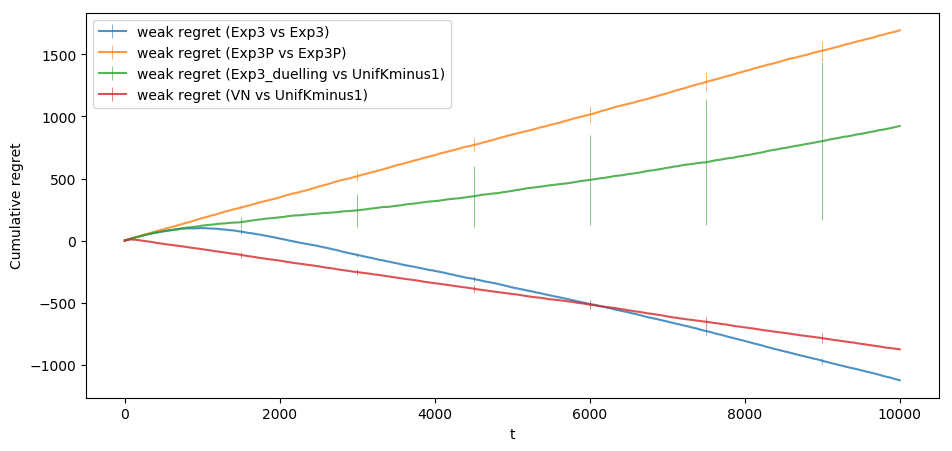}
\end{minipage}
\raggedright
\begin{minipage}{.5\textwidth}
  \centering
  \includegraphics[width=\textwidth]{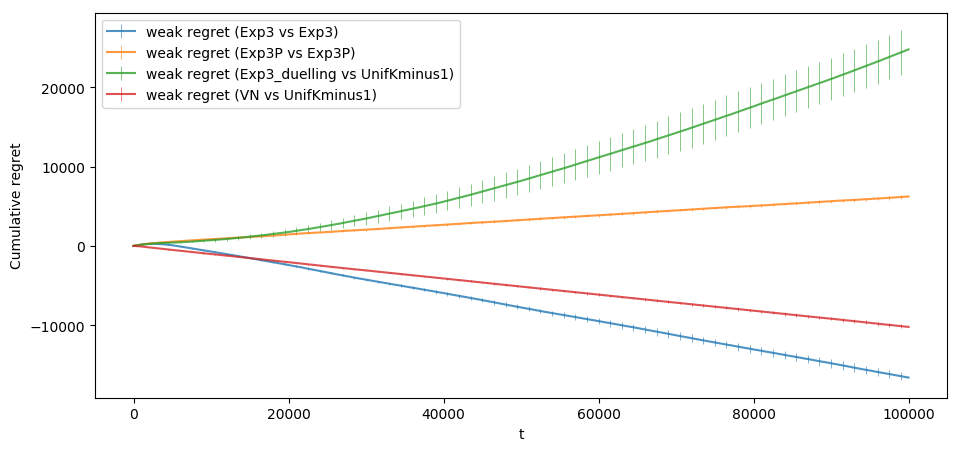}
\end{minipage}
\caption[Weak Regret of \nameref{alg:exp3_sparring},
\nameref{alg:exp3p_sparring}, \nameref{alg:exp3_unifkminus1}, and
\nameref{alg:vn_unifkminus1} in von-Neumann Setting]{Weak Regret of
  \nameref{alg:exp3_sparring}, \nameref{alg:exp3p_sparring},
\nameref{alg:exp3_unifkminus1}, and \nameref{alg:vn_unifkminus1} with
\(T=10^3\) (top-left), \(T=10^5\) (top-right), and \(T=10000\) (bottom-left)
in von-Neumann Setting}
  \label{fig:exp_vn}
\end{figure}

\section{Utility-Based Regret}
\label{sec:exp_util}
This section compares the empirical performance of
\nameref{alg:exp3_unifkminus1} to other algorithms in the stochastic
setting. Due to the absence of readily available
datasets inducing a linear order, we resort to a small synthetic preference
matrix. Again, we restrict ourselves to weak regret, as the expected
strong regret of \nameref{alg:exp3_unifkminus1} scales linearly in \(T\),
while the strong regret is an upper bound on the weak regret.

In addition to our modifications presented in Section~\ref{sec:exp_borda},
we implemented \texttt{REX3} according to \cite[Algorithm
1]{Gajane:2015:REW:3045118.3045143}, using \(\gamma = \min \left\{ \frac 1
2, \sqrt{\frac{K \log K}{0.5 T}} \right\}\).

\subsection{Experiment Setup}
We consider the following preference matrix, which induces a total order:
\begin{description}
  \item[Arithmetic] As suggested by \cite{pmlr-v40-Komiyama15}, this
    preference matrix assumes values \(P_{ij} = \frac 1 2 + \frac{j -
    i}{20}\) for all \(K = 8\) actions. This corresponds to a utility vector
    \(\mathbf x \in \left[ 0, 1 \right]^K\) with \(x_i = 1 - \frac i {10}\)
    and the linear link function~\eqref{eq:def_link_function_linear}.
\end{description}
All the simulation parameters are identical to the ones described in
Subsection~\ref{subsec:exp_borda_borda}, i.e. \(T = 10^4, 10^5\), we use 100
iterations, and all \texttt{UCB}-based algorithms use \(\alpha = 0.51\).

\subsection{Results}
\begin{figure}[H]
\begin{minipage}{.5\textwidth}
  \includegraphics[width=\textwidth]
  {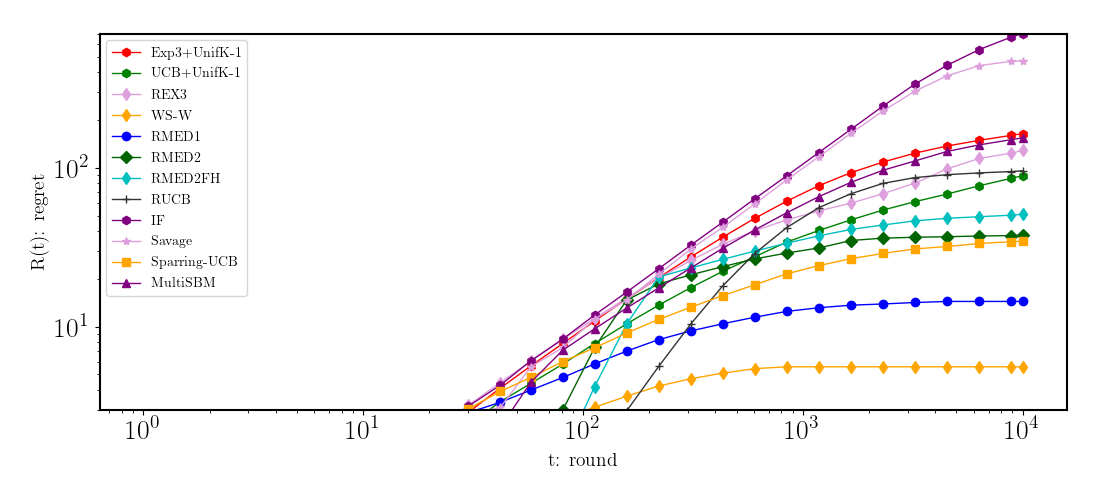}
\end{minipage}%
\begin{minipage}{.5\textwidth}
  \includegraphics[width=\textwidth]
  {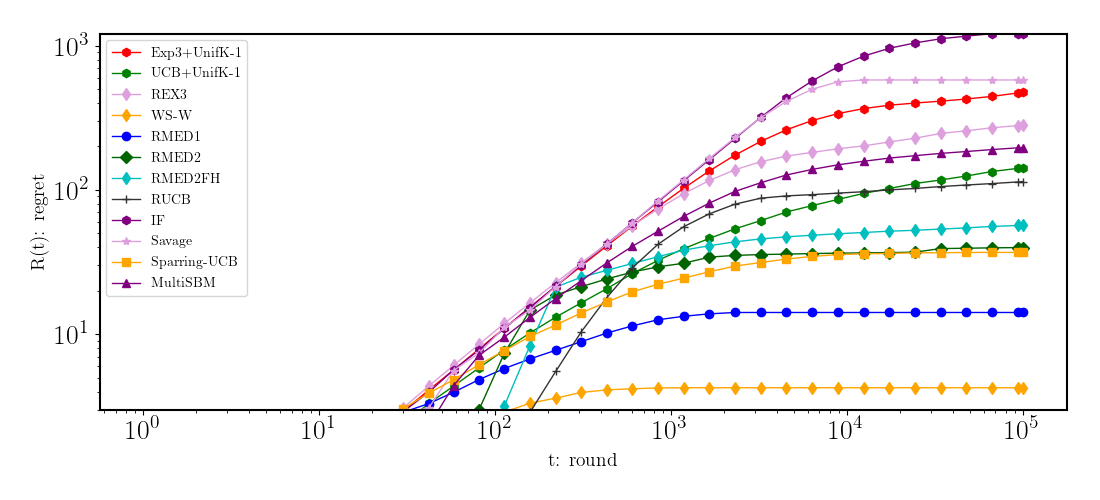}
\end{minipage}
\caption[Weak Regret of \nameref{alg:exp3_unifkminus1} in Stochastic
Utility-based Setting]{Weak regret of \nameref{alg:exp3_unifkminus1} and
other algorithms in the stochastic Utility-based setting with \(T=10^4\)
(left) and \(T=10^5\) (right) for \textbf{arithmetic}}
\label{fig:exp_utility_stochastic}
\end{figure}

Figure~\ref{fig:exp_utility_stochastic} shows the cumulative weak regret for
a range of algorithms. \nameref{alg:exp3_unifkminus1} behaves similarly as
in the Borda setting, suffering high regret in comparison to the more
specialised algorithms. \texttt{IF} suffers relatively
high regret, which is probably because of the small gap in preference
between the two best actions. As \texttt{SAVAGE}, \texttt{REX}, and
\nameref{alg:exp3_unifkminus1} are generic algorithms, they suffer higher
regret than their specialised counterparts \texttt{RMED} and \texttt{WS-W}.
\texttt{UCB+UnifK-1} performs slightly better than \texttt{REX3} and
\texttt{MultiSBM} in our experiments.

\chapter{Discussion}
\label{ch:discussion}
We have presented a duelling bandits algorithm designed for the adversarial
setting with the weak regret, effectively reducing the problem to a
classical bandit problem as suggested by previous literature
\citep{Urvoy:2013:GEK:3042817.3042904,
Zoghi:2014:RUC:3044805.3044894,pmlr-v38-jamieson15}. In the course of
proving upper bounds on the expected weak regret, we modified the
parameterisation, yielding a tighter upper bound in comparison with the
standard reduction. These bounds, which are comparable to those of
\texttt{Exp3}, hold for both the utility-based setting and the Borda winner
setting. The Condorcet setting, the Copeland setting, and
the von-Neumann setting can induce a winner which is different from the
Borda winner. As demonstrated experimentally, \nameref{alg:exp3_unifkminus1}
converges to the wrong action in these scenarios, accumulating regret linear
in
\(T\). While the Condorcet winner is indeed more appropriate than the Borda
winner in settings where the set of actions as a whole is less does not
directly influence an individual action's quality
\citep{Zoghi:2014:RUC:3044805.3044894}, there are scenarios where the
Borda winner has the properties of the desired winner. As the concept of the
Borda winner is rooted in voting theory, the Borda winner can be used to
resolve voting paradoxes when Condorcet-consistency is not required
\citep{Chevaleyre:2007:SIC:1417166.1417171}. \cite{pmlr-v38-jamieson15}
highlights the robustness of the Borda winner to estimation errors in the
preference matrix in contrast to the Condorcet winner. The argument of
low robustness is also applicable to the more general
Copeland winner. In conclusion, the answer to whether the
Borda winner is the appropriate modelling criterion depends solely on the
application's problem formulation.

When used in the stochastic setting, \nameref{alg:exp3_unifkminus1} and
\texttt{Exp3} share the same drawback of relatively high regret, as their
parameterisations effectively differ only by a multiplicative constant. In
case the
environment is known to be stochastic, our experiments suggested that
replacing the \texttt{Exp3} algorithm with a \texttt{UCB}-based algorithm
can reduce the regret. In the utility-based setting and the more general
Condorcet winner setting this modification is not enough to keep up with
more specialised algorithms such as \texttt{WS-W} and \texttt{RMED1}, whose
expected regret is upper bounded by \( O(\log T) \) or even \( O(1) \) in
the utility-based setting. The Borda setting has only been covered by
\cite{pmlr-v38-jamieson15} so far, who make some structural assumptions on
the preference matrix, allowing faster convergence to the winner than the
standard reduction scheme allows for. Applying algorithms with specific
performance guarantees in both the stochastic and the adversarial
environment, as presented by
\cite{pmlr-v23-bubeck12b,Seldin:2014:OPA:3044805.3045036}, is left for
future work.

For the sake of simplicity, we have assumed that the time horizon \(T\) is
known to the algorithm, allowing to optimise the choice of parameters. By
modifying the analysis, as performed by \cite{MAL-024},
\nameref{alg:exp3_unifkminus1} can be generalised to an anytime algorithm,
worsening the bound on the expected regret by a factor of \(\sqrt 2\).

Our theoretical analysis of \nameref{alg:exp3_unifkminus1} relies on the
linear link function. Generalisation to other link functions, e.g. the
Bradley-Terry model or the probit model, is left for future work.

Overall, \nameref{alg:exp3_unifkminus1} should be not be seen as an
algorithm optimised for weak regret, but a generic reduction to the cardinal
bandit problem with slightly tuned parameters. Due to its persistent uniform
exploration it is not suitable when considering strong regret. It is most
suitable for adversarial settings, where it is applicable to both the
utility-based setting and the Borda setting.

\renewcommand{\bibname}{References}
\bibliography{bib}{}
\bibliographystyle{plainnat}

\appendix
\chapter[Proof of Lemma 5.1]{Proof of Lemma~\ref{lemma:technical}}
\label{app:proof_lemma_technical}
\begin{proof}
  Let \(n \in \mathbb N, f : \mathbb R^n \to \mathbb R\) and
  \begin{equation*}
    f(\mathbf x) := -\sum_{i=1}^n \sum_{j=1}^n (x_i - x_j + 1)^2
    = 2 \sum_{i=1}^n \sum_{j=1}^n x_i x_j - 2n \sum_{i=1}^n x_i^2 - n^2
  \end{equation*}
  The above problem can be reformulated as constrained convex optimisation
  problem:
  \begin{equation*}
    \underset{\mathbf x \in \mathbb R^n}{\mathrm{minimise}}\,
    f(\mathbf x) \text{ such that } 0 \leq x_i \leq 1 \text{ for } i = 1,
    \hdots, n.
  \end{equation*}
  Using the set of functions
  \begin{align*}
    g_i(\mathbf x) &= x_i - 1 \\
    g_{n+i}(\mathbf x) &= -x_i,
  \end{align*}
  the two-sided inequality constraints can be expressed as
  \begin{equation*}
    g_i \leq 0 \text{ for } i = 1, \hdots, 2n,
    \label{eq:def_g_constraints}
  \end{equation*}
  yielding the following Lagrangian:
  \begin{align*}
    L(\mathbf x, \mathbf \lambda) &= f(\mathbf x) + \sum_{i=1}^{2n}
    \lambda_i g_i(\mathbf x) \\
    &= f(\mathbf x) + \sum_{i=1}^n \mathbf \lambda_i (x_i - 1) -
    \sum_{i=1}^n \lambda_{n+i} x_i.
  \end{align*}
  By complementary slackness, the solution \(\mathbf x^*, \mathbf
  \lambda^*\) satisfies
  \begin{equation}
    \lambda_i^* (x_i^* - 1) = \lambda_{n+i}^* x_i^* = 0
    \label{eq:complementary_slackness}
  \end{equation}
  for all \(i = 1, \hdots, n\). Let
  \(S_0 = \left\{ i \middle| x_i = 0 \right\},
    S_1 = \left\{ i \middle| x_i = 1 \right\},
    S_\lambda = \left\{ i \middle| 0 < x_i < 1 \right\}\) and
    \(n_0 = \left\lvert S_0 \right\rvert,
      n_1 = \left\lvert S_1 \right\rvert,
    n_\lambda = \left\lvert S_\lambda \right\rvert\). By
    \eqref{eq:complementary_slackness},
    \begin{equation*}
      \forall i \in S_\lambda: \lambda_i^* = \lambda_{n+i}^i = 0.
    \end{equation*}
    This simplifies the Lagrangian condition for all \(i \in S_\lambda\)
    to
    \begin{equation*}
      \frac{\partial L(\mathbf x^*, \mathbf \lambda^*)}{\partial x_i}
      = \frac{\partial f(\mathbf x^*)}{\partial x_i} = 0,
    \end{equation*}
    which is equivalent to
    \begin{equation*}
      \frac{\partial f(\mathbf x^*)}{\partial x_i}
      = 4 \sum_{j=1,j \neq i}^n x_j + 4 x_i - 4n x_i = 0.
    \end{equation*}
    Its solution is given by
    \begin{equation*}
      x_i = \frac 1 {n - 1} \sum_{j=1, j \neq i}^n x_j.
    \end{equation*}
    As
    \begin{align*}
      \sum_{j=1, j \neq i}^n x_j
      &= 0 n_0 + 1 n_1 + (n_\lambda - 1) x_i \\
      &= n_1 + (n_\lambda - 1) x_i,
    \end{align*}
    it follows that
    \begin{align*}
      x_i &= \frac{n_1}{(n - 1) \left( 1 - \frac{n_\lambda - 1}{n - 1}
      \right)} \\
      &= \frac{n_1}{n - n_\lambda} \\
      &= \frac{n_1}{n_0 + n_1}.
    \end{align*}
    As \(x_i \in [0, 1]\), \(n_0 + n_1 > 0\).  We can rewrite the original
    optimisation problem as integer linear programme
    \begin{align*}
      \underset{n_0, n_1, n_\lambda}{\mathrm{maximise}}\, &
      n^2 + 2 n_0 n_1 + 2 n_0 n_\lambda \left( \frac{n_1}{n_0 + n_1}
      \right)^2 + 2 n_1 n_\lambda \left( \frac{n_0}{n_0 + n_1} \right)^2
      \\ &= n^2 + 2 n_0 n_1  \left(1 + \frac{(n - (n_0 + n_1)) (n_0 +
      n_1)}{(n_0 + n_1)^2}\right) \\
        &= n^2 + \frac{2 n_0 n_1 n}{n_0 + n_1},
    \end{align*}
    such that \(n_0, n_1, n_\lambda \in \mathbb N_0, n_0 + n_1 +
    n_\lambda = n, n_0 + n_1 > 0\). By relaxing the last constraint with
    \(n_0 + n_1 \leq n\), the initial problem can be relaxed to the
    corresponding real-valued version
    \begin{equation*}
      n^2 + \frac{2 n_0 n_1 n}{n_0 + n_1} \leq \frac{3n^2} 2,
    \end{equation*}
    which is equivalent to
    \begin{equation}
      \frac{4 n_0 n_1}{n_0 + n_1} \leq n.
      \label{eq:lemma_technical_relaxed}
    \end{equation}
    We will now show that 
    \begin{equation}
      \frac{4 n_0 n_1}{n_0 + n_1} \leq n_0 + n_1.
      \label{eq:lemma_technical_step2}
    \end{equation}
    As \(n_0 + n_1 > 0\), this inequality can be rearranged to
    \begin{equation*}
      4 n_0 n_1 \leq (n_0 + n_1)^2,
    \end{equation*}
    which is equivalent to
    \begin{equation*}
      0 \leq (n_0 - n_1)^2,
    \end{equation*}
    proving \eqref{eq:lemma_technical_step2}, which in turn proves
    \eqref{eq:lemma_technical_relaxed}, as \(n_0 + n_1 \leq n\).
\end{proof}

\chapter{Experiments}
This appendix contains the scripts used to automate the experiments,
facilitating the reproduction of our results.

\section{\nameref{alg:exp3_unifkminus1} in the Stochastic Borda Setting}
\label{app:exp_stochastic_borda}
\begin{minipage}{\textwidth}
\begin{lstlisting}[caption=Experiment Setup for
\nameref{app:exp_stochastic_borda}]
#!/usr/bin/bash

set -e

base_filename="borda_stochastic_%s_%d"

git checkout borda
./compile
mkdir -p out plots

for pref in cyclic arxiv sushi; do
  for T in 10000 100000; do
    result_file="$(printf $base_filename.txt $pref $T)"
    plot_file="$(printf plots/$base_filename.png $pref $T)"
    build/main -r 100 -t "$T" -i prefmats/pref_${pref}.txt \
      -o "$result_file"
    python2 simpleplot.py "out/$result_file" "$plot_file"
  done
done

\end{lstlisting}
\end{minipage}

\section{\nameref{alg:exp3_unifkminus1}, \nameref{alg:exp3_sparring},
  \nameref{alg:exp3p_sparring}, \nameref{alg:vn_unifkminus1} in the Borda
Setting}
\label{app:exp_borda_vn}
\begin{minipage}{\textwidth}
\begin{lstlisting}[caption=Experiment Setup for \nameref{app:exp_borda_vn}]
#!/usr/bin/bash

base_filename="borda_vn_%d"

for rep in 50 500 5000; do
  T=$((rep * 20))
  seq_file="$(printf seq/$base_filename.xz $T)"
  result_file="$(printf results/$base_filename.xz $T)"
  plot_dir="$(printf plots/$base_filename $T)"
  
  ./gen_seq.py -o "$seq_file" -T 20 --rep "$rep" \
    --type gen_from_preference_matrix_rep -P prefs/borda_vonneumann

  ./run_simulation.py -i "$seq_file" --niter 100 -o "$result_file" \
    --algos Exp3+Exp3 Exp3P+Exp3P Exp3_duelling+UnifKminus1 VN+UnifKminus1

  ./plot_results.py -W Borda -i "$result_file" -o "$plot_dir"
done
\end{lstlisting}
\end{minipage}

\section{\nameref{alg:exp3_unifkminus1} in Copeland Setting}
\label{app:exp_copeland_exp3_unifkminus1}
\begin{minipage}{\textwidth}
\begin{lstlisting}[caption=Experiment Setup for
\nameref{app:exp_copeland_exp3_unifkminus1}]
#!/usr/bin/bash

base_filename="copeland_exp3_unifkminus1_%d"

for rep in 100 1000; do
  T=$((rep * 10))
  seq_file="$(printf seq/$base_filename.xz $T)"
  result_file="$(printf results/$base_filename.xz $T)"
  plot_dir="$(printf plots/$base_filename $T)"
  
  ./gen_seq.py -o "$seq_file" -T 10 --rep "$rep" \
    --type gen_from_preference_matrix_rep -P prefs/copeland_borda

  ./run_simulation.py -i "$seq_file" --niter 100 -o "$result_file" \
    --algos Exp3+UnifKminus1

  ./plot_results.py -W Copeland -i "$result_file" -o "$plot_dir"
done
\end{lstlisting}
\end{minipage}

\section{\nameref{alg:exp3_sparring}, \nameref{alg:exp3p_sparring},
\nameref{alg:vn_unifkminus1} in Copeland Setting}
\label{app:exp_copeland_vn}
\begin{minipage}{\textwidth}
\begin{lstlisting}[caption=Experiment Setup for \nameref{app:exp_copeland_vn}]
#!/usr/bin/bash

base_filename="copeland_vn_%d"

for rep in 25 250; do
  T=$((rep * 40))
  seq_file="$(printf seq/$base_filename.xz $T)"
  result_file="$(printf results/$base_filename.xz $T)"
  plot_dir="$(printf plots/$base_filename $T)"
  
  ./gen_seq.py -o "$seq_file" -T 40 --rep "$rep" \
    --type gen_from_preference_matrix_rep -P prefs/copeland_vonneumann

  ./run_simulation.py -i "$seq_file" --niter 100 -o "$result_file" \
    --algos Exp3+Exp3 Exp3P+Exp3P VN+UnifKminus1

  ./plot_results.py -W Copeland -i "$result_file" -o "$plot_dir"
done
\end{lstlisting}
\end{minipage}

\section{\nameref{alg:exp3_unifkminus1} in Von-Neumann Setting}
\label{app:exp_vn}
\begin{minipage}{\textwidth}
\begin{lstlisting}[caption=Experiment Setup for \nameref{app:exp_vn}]
#!/usr/bin/bash

base_filename="vn_%d"

for rep in 25 250 2500; do
  T=$((rep * 40))
  seq_file="$(printf seq/$base_filename.xz $T)"
  result_file="$(printf results/$base_filename.xz $T)"
  plot_dir="$(printf plots/$base_filename $T)"
  
  ./gen_seq.py -o "$seq_file" -T 40 --rep "$rep" \
    --type gen_from_preference_matrix_rep -P prefs/different_winners

  ./run_simulation.py -i "$seq_file" --niter 100 -o "$result_file" \
    --algos Exp3+Exp3 Exp3P+Exp3P Exp3_duelling+UnifKminus1 VN+UnifKminus1

  ./plot_results.py -W von-Neumann -i "$result_file" -o "$plot_dir"
done
\end{lstlisting}
\end{minipage}

\section{\nameref{alg:exp3_unifkminus1} in Stochastic Utility-based Setting}
\label{app:exp_stochastic_utility}
\begin{minipage}{\textwidth}
\begin{lstlisting}[caption=Experiment Setup for
\nameref{app:exp_stochastic_utility}]
#!/usr/bin/bash

set -e

base_filename="utility_stochastic_%s_%d"

git checkout utility
./compile
mkdir -p out plots

for pref in arithmetic; do
  for T in 10000 100000; do
    result_file="$(printf $base_filename.txt $pref $T)"
    plot_file="$(printf plots/$base_filename.png $pref $T)"
    build/main -r 100 -t "$T" -i prefmats/pref_${pref}.txt \
      -o "$result_file"
    python2 simpleplot.py "out/$result_file" "$plot_file"
  done
done

\end{lstlisting}
\end{minipage}

\end{document}